\documentclass[11pt]{article}

\usepackage[utf8]{inputenc} 
\usepackage[T1]{fontenc}    
\usepackage{hyperref}       
\usepackage{url}            
\usepackage{booktabs}       
\usepackage{amsfonts}       
\usepackage{textcomp}
\usepackage{csquotes}

\usepackage{tikz}
\usetikzlibrary{automata,arrows,patterns}
\usepackage{tango-colors}

\usepackage{pgfplots}
\pgfplotsset{compat=1.13}
\usepgfplotslibrary{groupplots}

\usepackage{algorithm}
\usepackage{algpseudocode}

\usepackage{amsmath}
\usepackage{amsthm}
\usepackage{amssymb}
\usepackage{bm}

\usepackage[a4paper, left=3cm, right=2.5cm, top=3cm, bottom=3cm]{geometry}

\usepackage{authblk}
\usepackage{natbib}
\usepackage{doi}

\newtheorem{thm}{Theorem}
\theoremstyle{definition}
\newtheorem{dfn}{Definition}

\newcommand{\N}{\mathbb{N}}
\newcommand{\R}{\mathbb{R}}
\newcommand{\aut}{\mathcal{A}}
\newcommand{\lng}{\mathcal{L}}
\newcommand{\res}{\mathcal{R}}

\newcommand{\trm}[1]{\textcolor{orange3}{#1}}
\newcommand{\nt}[1]{\textcolor{skyblue3}{#1}}

\renewcommand{\cite}{\citep}

\tikzstyle{vec}=[rounded corners, rectangle, semithick, minimum width=1.5cm]
\tikzstyle{inp}=[fill=skyblue1, draw=skyblue3, text=skyblue3]
\tikzstyle{stk}=[fill=orange1, draw=orange3, text=orange3]
\tikzstyle{nrn}=[fill=aluminium4, draw=aluminium6, text=aluminium6]

\tikzstyle{curves}=[line width=0.5mm]
\tikzstyle{curves_thin}=[line width=0.25mm]
\tikzstyle{curves_thick}=[line width=0.75mm]
\tikzstyle{class0color}=[aluminium6]
\tikzstyle{class0}=[draw=aluminium6, fill=aluminium4, text=aluminium6]
\tikzstyle{class1color}=[skyblue3]
\tikzstyle{class1}=[draw=skyblue3, fill=skyblue1, text=skyblue3]
\tikzstyle{class2color}=[orange3]
\tikzstyle{class2}=[draw=orange3, fill=orange1, text=orange3]
\tikzstyle{class3color}=[chameleon3]
\tikzstyle{class3}=[draw=chameleon3, fill=orange1, text=chameleon3]

\begin{document}
	
\title{Reservoir Stack Machines}

\author[1]{Benjamin Paaßen}
\author[2]{Alexander Schulz}
\author[2]{Barbara Hammer}
\affil[1]{Humboldt-University of Berlin}
\affil[2]{Bielefeld University}

\date{This is the accepted manuscript of \citet{Paassen2021} as provided by the authors. For the original version, refer to \doi{10.1016/j.neucom.2021.05.106}.
We share this manuscript according to the \href{https://www.elsevier.com/about/policies/sharing}{Elsevier guidelines on article sharing} under a \href{https://creativecommons.org/licenses/by-nc-nd/4.0/}{CC-BY-NC-ND} license.} 
\pagestyle{myheadings}
\markright{This is the accepted manuscript of \citet{Paassen2021} as provided by the authors.}

\maketitle

\begin{abstract}
Memory-augmented neural networks equip a recurrent neural network with an explicit memory
to support tasks that require information storage without interference over long times.
A key motivation for such research is to perform classic computation tasks, such as parsing. However,
memory-augmented neural networks are notoriously hard to train, requiring many backpropagation epochs and a lot of data.
In this paper, we introduce the reservoir stack machine, a model which can provably recognize all
deterministic context-free languages and circumvents the training problem by training only the
output layer of a recurrent net and employing auxiliary information during training about the
desired interaction with a stack. In our experiments, we validate the reservoir stack machine against deep and shallow
networks from the literature on three benchmark tasks for Neural Turing machines and
six deterministic context-free languages. Our results show that the reservoir stack machine
achieves zero error, even on test sequences longer than the training data, requiring only a few seconds of training time
and 100 training sequences.
\end{abstract}

\section{Introduction}

Memory-augmented neural networks (MANNs) are models that extend recurrent neural networks with an
explicit external memory \cite{NTM,MANN,MANNMeta}. A main motivation for such extensions is to
enable computation tasks that require a separation of memory and computation, i.e.\ the network
needs to keep track of information over long stretches of time without this information getting
corrupted in every step of the recurrent update process \cite{NTM}.
MANNs have achieved impressive successes on that front, performing computation
tasks that were previously out of reach for neural networks, such as associative memory recalls,
question-answering, and graph traversal \cite{DNCMDS,Pushdown,NTM,MANN,DeepStacks}.
Due to their ability to perform computation tasks, MANNs have also been named
Neural Turing Machines or Differentiable Neural Computers \cite{NTM}.

While the successes of this line of research are impressive, MANNs are typically hard to train,
requiring many epochs of gradient descent and a lot of training data \cite{NTM_impl}. This is because
one needs to backpropagate through a series of quasi-discrete read and write operations on the memory
with difficult dependencies. For example: accessing the memory is only useful if the memory contains
information that one needs to produce a certain output, and the memory only contains such information
if the network has previously put it into the memory.

In this paper, we introduce the \emph{reservoir stack machine} (RSM), an echo state network \cite{ESN}
combined with an explicit stack that can store information without interference but
is much simpler to train compared to the aforementioned MANN models thanks to two tricks.
First, we follow the reservoir computing paradigm
by leaving recurrent matrices fixed after initialization, limiting training to the output layers \cite{ESN,CRJ}.
Second, we do not require our network to learn optimal write/read behavior autonomously but provide
training data for it. Note that this is a restriction of our model because we need more
annotations per training sequence compared to standard recurrent neural networks.
However, this additional labeling makes training vastly more efficient:
Our training reduces to a classification
problem that can be solved with convex optimization in seconds instead of minutes to hours, using only
$\approx$ 100 short example inputs instead of hundreds of thousands.
One could see this approach as an instance of imitation learning, where learning
a complicated recurrent process becomes significantly simpler because we imitate the
actions of a teacher on a few example demonstrations \cite{Imitation}.

The RSM architecture is inspired by three main sources. First, we build on previous work regarding
differentiable stacks \cite{Pushdown,ExtDiffStacks,DeepStacks} which suggests a stack as memory
for neural networks to recognize typical context-free languages.
Second, we build on classic parser theory \cite{LR} to show that
the RSM is at least as powerful as an LR(1)-automata, i.e.\ it can recognize all 
deterministic context-free languages. Finally, our model builds upon prior work in reservoir computing,
which shows that echo state networks on their own have little computational power
\citep[below Chomsky-3;][]{DMM} but are well suited to distinguish input sequences by a constant-sized
suffix - which can be used to guide memory access behavior \cite{RMM2}.

In more detail, our contributions are as follows.
\begin{itemize}
	\item We introduce a novel MANN architecture, namely the reservoir stack
	machine (RSM), which combines an echo state network \cite{ESN} with an explicit stack to store inputs as
	well as auxiliary symbols.
	\item We prove that RSMs are at least as powerful as LR(1)-automata and thus can recognize
	all deterministic context-free languages \cite{LR}, whereas past reservoir memory machines \cite{RMM2}
	cannot.
	\item We evaluate our model on three benchmarks for Neural Turing Machines (latch, copy, and
	repeat copy) and on six context-free languages, verifying that it can learn all the tasks in a few
	seconds of training time from only $100$ training sequences. By contrast, we also show that
	deep models (namely GRUs \cite{GRU}, the deep stack model of Suzgun et al. \cite{DeepStacks}, and
	an end-to-end trained version of our proposed model)
	need much longer training times and sometimes fail to learn the task.
\end{itemize}

We note that our contributions are largely conceptual/theoretical. We do not expect
that our model is widely applicable to practical tasks because required teaching signals for
the dynamics might be missing. Instead, we investigate the interface between theoretical computer
science and reservoir computing, providing further insight into the computational capabilities of
neural networks, especially reservoir neural networks.

We begin our paper by describing background knowledge as well as related work
(refer to Section~\ref{sec:background}), then introducing our model (refer to Section~\ref{sec:method}),
and finally describing the experimental evaluation (refer to Section~\ref{sec:experiments}).

\section{Background and Related Work}
\label{sec:background}

Our work in this paper is connected to several established lines of research. Due to
the breadth of connected fields, we can only touch upon key ideas. Readers are invited to follow
our references to more detailed discussions of the underlying topics. We begin with the
central topic of our investigation, namely the computational power of neural networks,
connect this to grammatical inference and memory-augmented neural nets, before
introducing the underlying concepts if our reservoir stack machine model, namely LR(1)
automata and echo state networks.

\subsection{Computational Power of Recurrent Neural Networks}

Computational power is typically measured by comparing a computational system to a reference
automaton class in the Chomsky hierarchy, that is finite state automata (Chomsky-3), LR automata
(Chomsky-2), linearly space-bounded Turing machines (Chomsky-1), or Turing machines (Chomsky-0)
\cite{TheoInf}. For the sake of space we do not define every automaton class in detail.
We merely remark that, for our purpose, any automaton implements a function $\aut :
\Sigma^* \to \{0, 1\}$ mapping input sequences $\bar x$ over some set $\Sigma$ to zero or one.
If $\aut(\bar x) = 1$ we say that the automaton \emph{accepts} the word. Otherwise, we
say that it \emph{rejects} the word. Accordingly, we define the \emph{language} $\mathcal{L}(\aut)$ 
accepted by the automaton as the set $\mathcal{L}(\aut) := \{ \bar x \in \Sigma^* |
\aut(\bar x) = 1 \}$.

In this paper, we are interested in the computational power of neural networks.
We particularly focus on recurrent neural networks (RNNs), which we define as follows.

\begin{dfn}[Recurrent neural network]
	A \emph{recurrent neural network} with $n$ inputs, $m$ neurons, and $k$ outputs is defined as
	a $4$-tuple $(\bm{U}, \bm{W}, \sigma, g)$, where $\bm{U} \in \R^{m \times n}$, $\bm{W} \in \R^{m \times m}$,
	$\sigma : \R \to \R$, and $g : \R^m \to \R^k$.
	
	Now, let $\vec x_1, \ldots, \vec x_T$ be a sequence with $T$ elements over $\R^n$.
	We define the \emph{state} $h_t \in \R^m$ and the output $y_t \in \R^k$ at time $t$ via the equations:
	\begin{align}
		\vec h_t &= \sigma\Big(\bm{U} \cdot \vec x_t + \bm{W} \cdot \vec h_{t-1}\Big), \label{eq:rnn} \\
		\vec y_t &= g(\vec h_t),
	\end{align}
	where $\sigma$ is applied element-wise and $\vec h_0 = \vec 0$.
\end{dfn}

Now, let $\Sigma \subset \R^n$ be some set and $\aut : \Sigma^* \to \{0,1\}$ be the
function implemented by an automaton.
We say that an RNN \emph{simulates} the automaton if for
any input sequence $\vec x_1, \ldots, \vec x_T \in \Sigma^*$ we obtain
$\vec y_T = \aut(\vec x_1, \ldots, \vec x_T)$. Early work has already demonstrated
that recurrent neural networks (RNNs) with integer weights are sufficient to simulate finite state
automata \cite{FiniteStateNets}, that RNNs with rational weights can simulate
Turing machines \cite{RecTuring}, and that RNNs with real weights
even permit super-Turing computations \cite{SuperTuring}. Recently, Šíma has shown that
binary RNNs with 1 or 2 extra rational-valued neurons lie in between
Chomsky-3 and Chomsky-0, each recognizing some but not all languages in the in-between classes
\cite{AnalogNeurons}.

Importantly, all these results rely on deterministic constructions to transform a known automaton
into a neural network with carefully chosen weights. Learning languages from examples is a much
more difficult task, studied under the umbrella of grammatical inference.

\subsection{Grammatical inference}

Grammatical inference (or grammar induction) is concerned with inferring a parser for a language
just from examples for words that should be accepted (positive examples) and/or words that should be
rejected (negative examples) \cite{GrammarInference}. In general, this is a hard problem. For example, 
even some Chomsky-3 languages can not be learned from positive examples alone \cite{GrammarInference}.
Further, grammar inference is typically ambiguous, i.e.\ there may be infinitely many parsers which
recognize the same language - and selecting the 'smallest' among them is typically NP-hard
\cite{GrammarInference,TheoInf}. Despite these difficulties, impressive progress has been made.
For example, the Omphalos challenge has sparked new research into learning deterministic
context-free grammars (DCFGs) \cite{Omphalos}. \cite{CPCFG,PACCFG} have shown that special subclasses
of probabilistic context-free grammars are learnable from examples.
Finally, \cite{RNNG_old,RNNG,VTBP,OrderedNeurons,ComposeWords} have suggested
neural models to learn grammatical structure from data. Our work in this paper is less general than
grammar inference approaches because we require training data on the desired
parsing behavior for (short) examples instead of learning this behavior from scratch. However,
our approach is more general than the deterministic translation schemes mentioned in the previous
section because we do not require a full parser specification - only its behavior on a 
set of training data. As such, our approach lies between constructive proofs of computational 
power and grammatical inference. In a sense, our scenario resembles imitation learning, where
a teacher demonstrates the correct (or at least viable) actions on a few training data instances
and the models task is to generalize this behavior to the entire space of possible inputs
\cite{Imitation}.

Most importantly, though, our focus does not lie on inferring grammatical structure, but, rather,
on studying the effect of a memory augmentation on computational power.

\subsection{Memory-Augmented Neural Networks}

For the purpose of this paper, we define a memory-augmented neural network (MANN) as any
neural network that is extended with an explicit write and read operation to interact with an
external memory. Recently, such models have received heightened interest, triggered by the
Neural Turing Machine \cite{NTM}. The Neural Turing Machine implements a content-based addressing 
scheme, i.e.\ memory addresses are selected by comparing content in memory to a query vector
and assigning attention based on similarity. The memory read is then given as the average of all
memory lines, weighted by attention.
Writing occurs with a similar mechanism. To support location-based addressing, the
authors introduce the concept of a linkage matrix which keeps track of the order in which content
was written to memory such that the network can decide to access whichever content was written 
after or before a queried line in memory. Further research has revised the model with
sparser memory access \cite{MANN}, refined sharpening operations \cite{DNCMDS}, or more efficient
initialization schemes \cite{NTM_impl}. However, it remains that MANNs are difficult to train,
often requiring tens of thousands of training epochs to converge even for simple memory access 
tasks \cite{NTM_impl}. In this paper, we try to simplify this training by employing echo state networks
(see below).

More precisely, we focus on a certain kind of MANN, namely stack-based models.
Interestingly, such models go back far longer than the Neural Turing Machine, at least as far as
the Neural State Pushdown Automaton of \citet{Pushdown}. Such a model
involves differentiable push and pop operations to write content onto the stack and remove it again.
Recently, Suzgun et al.\ have built upon this concept and introduced stack recurrent neural 
networks \cite{DeepStacks} which can learn typical context free languages.
A stack model is particularly appealing because it reduces the interaction with memory to two 
very basic operations, namely to push one single piece of information onto the stack, and to pop 
one single piece of information from it - no sharpening operations or linking matrices required. 
Further, despite its simplicity, a stack is sufficient to implement LR(1)-automata,
which we discuss next.

\subsection{LR(1)-automata and the computational power of stacks}

We define an LR(1)-automaton as follows.

\begin{dfn}
An \emph{LR(1)-automaton} is defined as a $4$-tuple $\aut = (\Phi, \Sigma, R, F)$,
where $\Phi$ and $\Sigma$ are both finite sets with $\Phi \cap \Sigma = \emptyset$,
where $F \subseteq \Phi$, and where $R$ is a list of $4$-tuples of the form $(\bar s, x, j, A)$
with $\bar s \in (\Phi \cup \Sigma)^*$, $x \in \Sigma \cup \{\varepsilon\}$, $j \in \N_0$,
and $A \in \Phi$. $\varepsilon$ denotes the empty word. We call $\Phi$ the nonterminal symbols, $\Sigma$ the terminal symbols,
$R$ the rules, and $F$ the accepting nonterminal symbols of the automaton.

In a slight abuse of notation, we also use the symbol $\aut$ to denote the function
$\aut : \Sigma^* \to \{0, 1\}$ of the automaton, which we define as the result of
Algorithm~\ref{alg:parse} for the automaton $\aut$ and the input $\bar w \in \Sigma^*$.
We define the \emph{accepted language} $\lng(\aut)$ of $\aut$ as the set
$\lng(\aut) = \{ \bar w \in \Sigma^* | \aut(\bar w) = 1\}$.
Finally, we define the set LR(1) as the set of all languages $\lng$ that can be accepted
by some LR(1)-automaton.
\end{dfn}

\begin{algorithm}
\caption{The parsing algorithm for LR(1)-automata. The $\exists$ quantifier in line 4 always
chooses the first matching rule in the list $R$.}
\label{alg:parse}
\begin{algorithmic}[1]
\Function{parse}{automaton $\aut = (\Phi, \Sigma, R, F)$, word $w_1, \ldots, w_T \in \Sigma^*$}
\State Initialize an empty stack $\mathcal{S}$.
\For{$y \gets w_1, \ldots, w_T, \#$}
\While{$\exists \, (\bar s, x, j, A) \in R:$ $\bar s$ is suffix of $\mathcal{S}$ and $x \in \{ y, \varepsilon\}$}
\State Pop $j$ elements from $\mathcal{S}$.
\State Push $A$ onto $\mathcal{S}$.
\EndWhile
\State Push $y$ onto $\mathcal{S}$.
\EndFor
\If{$\exists A \in F : \mathcal{S} = A\#$}
\State \Return $1$.
\Else
\State \Return $0$.
\EndIf
\EndFunction
\end{algorithmic}
\end{algorithm}

Algorithm~\ref{alg:parse} works as follows: In each iteration, we consider one letter $y$ of the
input. Then, we check all rules $(\bar s, x, j, A) \in R$ in order.
In particular, we check if $\bar s$ is a suffix of the current stack
and if $x$ is either $y$ or $\varepsilon$. If both conditions hold, we say that the rule \emph{matches}.
In that case, we pop the top $j$ elements from
the stack and instead push the nonterminal $A$ onto the stack. Once no rule matches anymore,
we push $y$ onto $S$ and continue. The last letter we process is a special end symbol $\#$.
If only $A\#$ is left on the stack for some nonterminal $A \in F$, we accept the word and return $1$. 
Otherwise, we reject the word and return $0$.

As an example, consider the language $\lng_{a^nb^n} := \{ab, aabb, aaabbb, \ldots\}$.
This language is accepted by following LR(1)-automaton:
\begin{equation}
\aut_{a^nb^n} = \Big(\{S\}, \{a, b\}, [(aSb, \varepsilon, 3, S), (a, b, 0, S)], \{S\}\Big),
\label{eq:anbn}
\end{equation}
where $[]$ denotes an ordered list.
Now, consider the word $aabb$. When we apply Algorithm~\ref{alg:parse}, we notice that
no rule matches for the first and second input letter. Accordingly, the first three states of the
stack are $\varepsilon$ (the empty stack), $a$, and $aa$. At this point, the current input symbol
is $y = b$. Now, the rule $(a, b, 0, S)$ matches and, hence, the next stack state is
$aaS$. Since no rule matches anymore, we proceed and obtain the next stack state $aaSb$. Now, the
rule $(aSb, \varepsilon, 3, S)$ matches and we obtain the stack state $aS$. Again, no rule matches
anymore and we proceed to $aSb$. Now, rule $(aSb, \varepsilon, 3, S)$ matches again and we obtain
the stack $S$ and the loop in lines 3-9 completes with the stack $S\#$.
Finally, because $S \in F$, we obtain the output $1$.

It is well known that the computational power of LR(1)-automata lies between
Chomsky-3 and Chomsky-2 and corresponds exactly to deterministic context-free
languages.

\begin{thm}
Chomsky-3 $\subsetneq$ LR(1) $=$ DCFG $\subsetneq$ Chomsky-2.
\end{thm}
For a proof, refer to standard text books such as \citet{TheoInf}.

Interestingly, if we provide a model with a second stack, this is sufficient to simulate an
entire Turing machine, raising the computational power to Chomsky-0 \cite{TheoInf,DeepStacks}.
Overall, stacks appear as a particularly promising datastructure to augment neural nets with
more computational power. In this paper, we augment a neural net with a parsing
scheme similar to Algorithm~\ref{alg:parse}. The underlying neural network is an echo state network.

\subsection{Echo State Networks}

We define an echo state network as follows.

\begin{dfn}[Reservoir and echo state network]\label{dfn:esn}
We define a \emph{reservoir} with $n$ inputs and $m$ neurons as a triple
$\res = (\bm{U}, \bm{W}, \sigma)$ where $\bm{U} \in \R^{m \times n}$,
$\bm{W} \in \R^{m \times m}$, and $\sigma : \R \to \R$.

We say that a reservoir is \emph{compact} with respect to a compact set
$\Sigma \subset \R^n$ if there exists a compact set $\mathcal{H} \subset \R^m$
such that $\vec 0 \in \mathcal{H}$ and for all $\vec h \in \mathcal{H}$ as well as all
$\vec x \in \Sigma$ it holds: $\sigma(\bm{W} \cdot \vec h + \bm{U} \cdot \vec x)
\in \mathcal{H}$.

We further say that a reservoir fulfills the \emph{echo state property}
with respect to a compact set $\Sigma \subset \R^n$ if it is compact with respect to $\Sigma$ and
there exists an infinite null-sequence $\delta_1, \delta_2, \ldots$, such that for any infinite
input sequence $\vec x_1, \vec x_2, \ldots$ over $\Sigma$ and any two initial states $\vec h_0, \vec h'_0 \in \mathcal{H}$,
it holds: $\lVert \vec h_t - \vec h'_t \rVert \leq \delta_t$,
where $\vec h_t = \sigma(\bm{W} \cdot \vec h_{t-1} + \bm{U} \cdot \vec x_t)$
and $\vec h'_t = \sigma(\bm{W} \cdot \vec h'_{t-1} + \bm{U} \cdot \vec x_t)$
for all $t \in \N$.

Now, let $\Sigma \subset \R^n$ be a compact set.
We define an \emph{echo state network} (ESN) over $\Sigma$ with $m$ neurons as a recurrent
neural network $(\bm{U}, \bm{W}, \sigma, g)$ where $(\bm{U}, \bm{W}, \sigma)$
is a reservoir  with $n$ inputs and $m$ neurons that fulfills the echo state property with
respect to $\Sigma$.
\end{dfn}

Note that our definition of the echo state property slightly deviates from the original definition,
but is provably equivalent \cite{EchoStateProperty}.

Intuitively, the echo state property guarantees that the state $\vec h_t$
of a reservoir represents the  past via a fractal-like encoding that is dominated by its
suffix \cite{FractalESN}. This ensures that the reservoir reacts similarly to all
sequences that share the same suffix and thus can be viewed as a Markovian filter \cite{MarkovESN}.
Importantly, the specific parameters of the network do not matter much. As long as the echo
state property holds, the reservoir state provides a representation of the recent past and, thus,
we only need to adapt the output function $g$ to the task at hand, without any change to the reservoir.
This makes echo state networks appealing, because $g$ can usually be trained quickly using convex
optimization, whereas the adaptation of $\bm{U}$ and $\bm{W}$ requires heuristic schemes \cite{ESN}.
This is also our main motivation for using reservoir computing: They are a promising starting point
for networks that provide computational capabilities but remain fast to train.

However, the echo state property also limits the computational power of ESNs to languages
that can be recognized by a finite suffix \citep[i.e.\ definite memory machines;][]{DMM}.
For example, consider the language $ab^*$. For any $T \in \N$, this language contains the string $ab^T$ 
but not the string $b^T$. However, the echo state property enforces that the reservoir states
for these two strings become eventually indistinguishable, meaning that no (uniformly continuous)
output function $g$ can map them to different values.
It is currently an open question whether echo state networks that operate at the
edge of chaos extend this capability \cite{EdgeChaos,MemCap,MarkovESN,DeepESN}.
The provable computational limitations of ESNs make them a particularly interesting object of study
for memory-augmented neural networks because any computational capability beyond the limits of
definite memory machines \cite{DMM} must provably be due to the augmentation.

We note that our prior work has already investigated the computational effect of some
memory augmentations to ESNs.
In particular, this paper is an extension of the ESANN 2020 Paper
\enquote{Reservoir Memory Machines} \cite{RMM}. In contrast to this prior work, we do
not use a constant-sized memory but instead a stack, and we prove that this stack
improves computational power to that of LR(1) automata. In a separate and distinct
extension \cite{RMM2}, we kept the constant-sized memory but simplified the memory access
behavior and added an associative memory access mechanism. Such networks can provably simulate
any finite state automaton. However, we prove later in Section~\ref{sec:theory} that this memory
does not suffice to simulate general LR(1) automata. Instead, we use a stack as memory architecture
and prove that such a network can simulate all LR(1) automata.
Finally, the experimental evaluation for both papers are mostly disjoint, with \citet{RMM2}
focusing on associative memory and recall tasks as well as
finite state machines, whereas, in this paper, we focus on LR(1) language recognition.

\section{Method}
\label{sec:method}

In this section, we introduce our proposed model, the reservoir stack machine. First, we
define the dynamics of our model and explain its intended mechanism. Then, in Section~\ref{sec:theory},
we prove that a reservoir stack machine with a sufficiently rich reservoir can simulate any
LR(1)-automaton, while the reservoir memory machine \cite{RMM2} cannot.
Finally, we provide a training scheme in Section~\ref{sec:training}.

\begin{figure}
\begin{center}
\begin{tikzpicture}
\begin{scope}[shift={(0,+0.6)}]
\node[vec, inp] at (0,0) {$\vec x_1$};
\node[skyblue3] (inp) at (0,0.7) {$\vdots$};
\node[vec, inp] at (0,1.3) {$\vec x_t$};
\end{scope}

\begin{scope}[shift={(0,-0.6)}]
\node[vec, stk] at (0,-1.3) {$\vec s_1$};
\node[orange3]  (stk) at (0,-0.6) {$\vdots$};
\node[vec, stk] at (0,0) {$\vec s_\tau$};
\end{scope}

\begin{scope}[shift={(3,0)}]
\node[circle, nrn] (h1) at (-36:1.2) {};
\node[circle, nrn] (h2) at (-108:1.2) {};
\node[circle, nrn] (h3) at (-180:1.2) {};
\node[circle, nrn] (h4) at (-252:1.2) {};
\node[circle, nrn] (h5) at (-324:1.2) {};

\node at (0,0) {$\bm{W}$};

\path[nrn, ->, >=stealth', semithick]
(h1) edge[bend left] (h2)
(h2) edge[bend left] (h3)
(h3) edge[bend left] (h4)
(h4) edge[bend left] (h5)
(h5) edge[bend left] (h1);
\path[nrn, <->, >=stealth', semithick]
(h1) edge (h3)
(h3) edge (h5)
(h5) edge (h2)
(h2) edge (h4)
(h4) edge (h1);

\node at (-1.8,0) {$\bm{U}$};

\node (inp_to_res) at (135:1.2) {};
\node (stk_to_res) at (225:1.2) {};
\node (res_to_h)   at (45:1.2) {};
\node (res_to_g)   at (-45:1.2) {};

\path[inp, ->, >=stealth', semithick]
(inp) edge[out=0,in=135, shorten <=0.5cm] (inp_to_res);

\path[stk, ->, >=stealth', semithick]
(stk) edge[out=0,in=225, shorten <=0.5cm] (stk_to_res);
\end{scope}

\begin{scope}[shift={(5.5,0)}]

\node[vec, inp] (ht) at (0,+0.4)  {$\strut\vec h_t$};
\path[inp, ->, >=stealth', semithick]
(res_to_h) edge[out=45,in=180] (ht);

\node[vec, stk] (gt) at (0,-0.4) {$\strut\vec g_t$};
\path[stk, ->, >=stealth', semithick]
(res_to_g) edge[out=-45,in=180] (gt);

\node (state) at (0.6,0) {};

\end{scope}

\begin{scope}[shift={(7.6,0.5)}]

\node[right, aluminium6] (cpop)   at (0,-0.4) {$c^\text{pop}$};

\node[right, orange3] (j) at (1.5,-0.4) {$j$};

\node[right, aluminium6] (cpush)  at (0,+0.4) {$c^\text{push}$};

\node[right, orange3] (k) at (1.5,+0.4) {$\vec a$};

\node[right, aluminium6] (cshift) at (0,+1.2) {$c^\text{shift}$};

\node[right, orange3] (shift)  at (1.5,+1.2) {$0/1$};

\node[right, aluminium6] (cout)   at (0,-1.2) {$c^\text{out}$};

\node[vec, nrn] (yt) at (0.4,-2.35) {$\vec y_t$};

\path[nrn, ->, >=stealth', semithick]
(state)  edge[out=0,in=180] (cpop)
(cpop)   edge (j)
(state)  edge[out=0,in=180] (cpush)
(cpush)  edge (k)
(state)  edge[out=0,in=180] (cshift)
(cshift) edge (shift)
(state)  edge[out=0,in=180] (cout)
(cout)   edge (yt);

\end{scope}

\begin{scope}[shift={(11.5,-1.2)}]
\node[vec, stk]         at (0,0)   {$\vec s_1$};
\node[orange3]          at (0,0.7) {$\vdots$};
\node[vec, stk] (stop)  at (0,1.3) {$\vec s_{\tau-j}$};
\node[vec, stk] (snont) at (0,2.1) {$\vec a$};
\node[vec, stk] (sinp)  at (0,2.9) {$\vec x_t$};

\path[stk, ->, >=stealth', semithick]
(shift) edge (sinp)
(k)     edge (snont)
(j)     edge (stop);
\end{scope}

\end{tikzpicture}
\end{center}
\caption{An illustration of the reservoir stack machine dynamics.
The input sequence up to time $t$ is represented as a state $\vec h_t$
via the reservoir $(\bm{U}, \bm{W}, \sigma)$ and the current stack is represented
as a state $\vec g_t$ via the same reservoir. The concatenated state
$(\vec h_t, \vec g_t)$ is plugged into classifiers $c^\text{shift}$, $c^\text{push}$, and
$c^\text{pop}$, which generate the next stack state (right). Note that
$c^\text{pop}$ and $c^\text{push}$ are called repeatedly until they both return $0$.
The output function $c^\text{out}$ generates the next output $\vec y_t$.}
\label{fig:rsm}
\end{figure}
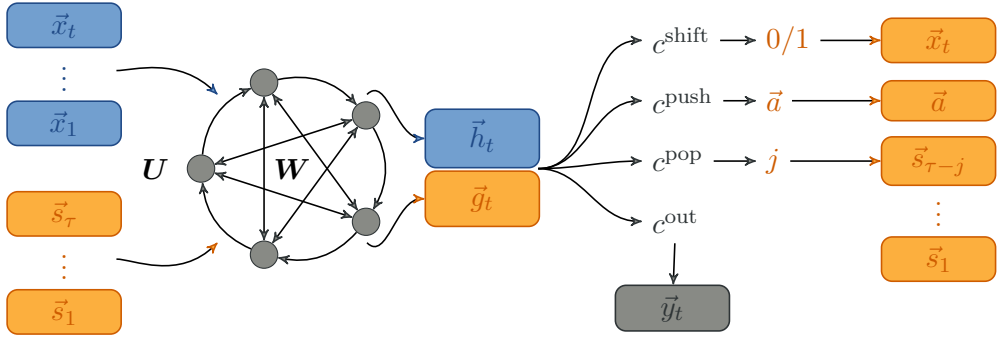

We construct a reservoir stack machine (RSM) as a combination of an echo state
network (ESN) \cite{ESN} with a stack, where the stack is controlled similarly to an LR(1)-automaton
\cite{LR}. In more detail:

\begin{dfn}
Let $\Sigma \subset \R^n$ and $\Phi \subset \R^n$ be compact sets with $\Sigma \cap \Phi = \emptyset$
and $\vec 0 \notin \Phi$.
We define a reservoir stack machine (RSM) over $\Sigma$ and $\Phi$ with $m$ neurons, $J$
maximally popped symbols, and $L$ outputs as a $7$-tuple
$\mathcal{M} = (\bm{U}, \bm{W}, \sigma, c^\text{pop}, c^\text{push},$ $c^\text{shift}, c^\text{out})$
where $(\bm{U}, \bm{W}, \sigma)$ is a reservoir with $n$ inputs and $m$ neurons that
conforms to the echo state property on $\Sigma \cup \Phi$, and where
$c^\text{pop} : \R^{2m} \to \{0, \ldots, J\}$, 
$c^\text{push} : \R^{2m} \to \Phi \cup \{\vec 0\}$,
$c^\text{shift} : \R^{2m} \to \{0, 1\}$, and
$c^\text{out} : \R^{2m} \to \R^L$.

Let $\vec x_1, \ldots, \vec x_T$ be a sequence over $\Sigma$. We define the output sequence
$\mathcal{M}(\vec x_1, \ldots, \vec x_T) = \vec y_1, \ldots, \vec y_{T+1}$
as the output of Algorithm~\ref{alg:rsm} for the input $\vec x_1, \ldots, \vec x_T$.
We say that an RSM simulates an automaton $\aut$ if $y_{T+1} = \aut(\vec x_1, \ldots, \vec x_T)$
for any input sequence $\vec x_1, \ldots, \vec x_T \in \Sigma^*$.
\end{dfn}

\begin{algorithm}
\caption{The dynamics of a reservoir stack machine
$\mathcal{M} = (\bm{U}, \bm{W}, \sigma, c^\text{pop}, c^\text{push}, c^\text{shift}, c^\text{out})$
on the input sequence $\vec x_1, \ldots, \vec x_T$.}
\label{alg:rsm}
\begin{algorithmic}[1]
\Function{RSM}{Input sequence $\vec x_1, \ldots, \vec x_T$}
\State Set $\vec x_{T+1} \gets \vec 0$.
\State Initialize an empty stack $\mathcal{S}$.
\State Initialize $\vec h_0 \gets \vec 0$, $\vec g_1 \gets \vec 0$.
\For{$t \gets 1, \ldots, T+1$}
\State $\vec h_t \gets \sigma\Big(\bm{U} \cdot \vec x_t + \bm{W} \cdot \vec h_{t-1}\Big)$.
\While{True}
\State $j \gets c^\text{pop}(\vec h_t, \vec g_t)$.
\If{$j > 0$}
\State Pop $j$ elements from $\mathcal{S}$.
\State $\vec g_t \gets \vec 0$.
\For{$\vec s \gets$ elements of $\mathcal{S}$}
\State $\vec g_t \gets \sigma\Big(\bm{U} \cdot \vec s + \bm{W} \cdot \vec g_t \Big)$.
\EndFor
\EndIf
\State $\vec a \gets c^\text{push}(\vec h_t, \vec g_t)$.
\If{$\vec a \neq \vec 0$}
\State Push $\vec a$ onto $\mathcal{S}$.
\State $\vec g_t \gets \sigma\Big(\bm{U} \cdot \vec a + \bm{W} \cdot \vec g_t \Big)$.
\EndIf
\If{$j = 0$ and $\vec a = \vec 0$}
\State \textbf{Break}.
\EndIf
\EndWhile
\State $\vec y_t \gets c^{\text{out}}(\vec h_t, \vec g_t)$.
\If{$c^\text{shift}(\vec h_t, \vec g_t) > 0$}
\State Push $\vec x_t$ onto $\mathcal{S}$.
\State $\vec g_{t+1} \gets \sigma\Big(\bm{U} \cdot \vec x_t + \bm{W} \cdot \vec g_t \Big)$.
\Else
\State $\vec g_{t+1} \gets \vec g_t$.
\EndIf
\EndFor
\State \Return $\vec y_1, \ldots, \vec y_{T+1}$.
\EndFunction
\end{algorithmic}
\end{algorithm}

Roughly speaking, Algorithm~\ref{alg:rsm} works as follows.
In every iteration $t$ of the RSM, we represent the input sequence
$\vec x_1, \ldots, \vec x_t$ up to time $t$ with a state vector $\vec h_t$, and the stack content
$\vec s_1, \ldots, \vec s_\tau$ with a state vector $\vec g_t$, both using the
same reservoir $(\bm{U}, \bm{W}, \sigma)$. Based on the concatenated
state vector $(\vec h_t, \vec g_t)$, we then let $c^\text{pop}$ decide how many symbols to pop from the
stack and we let $c^\text{push}$ decide which symbol (if any) to push on the stack until both classifiers
return $0$. Then, we let a function
$c^\text{out}$ decide the current output $\vec y_t$, and we let a binary classifier
$c^\text{shift}$ decide whether to push the current input symbol $\vec x_t$ onto the stack or not.
Every time we change the stack, we update the state $\vec g_t$ accordingly. Refer to Figure~\ref{fig:rsm}
for a graphical illustration.

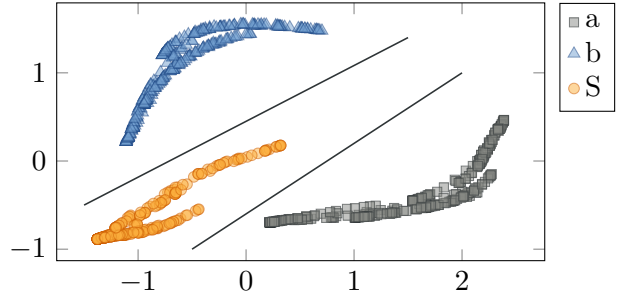
\begin{figure}
\begin{center}
\begin{tikzpicture}
\begin{axis}[width=8cm, height=5cm, legend pos={outer north east}, legend cell align=left]
\addplot[scatter,only marks,
scatter src=explicit,
scatter/classes={
0={mark=square*,fill=aluminium4,draw=aluminium6,mark size=0.06cm,opacity=0.5},%
1={mark=triangle*,fill=skyblue1,draw=skyblue3,mark size=0.09cm,opacity=0.5},%
2={mark=*,fill=orange1,draw=orange3,opacity=0.5}}]
table[x=x,y=y,meta=rule,col sep=tab]{reservoir_example.csv};
\legend{a, b, S}
\draw[semithick, aluminium6] (axis cs:-0.5,-1) -- (axis cs:2,1);
\draw[semithick, aluminium6] (axis cs:-1.5,-0.5) -- (axis cs:1.5,1.4);
\end{axis}
\end{tikzpicture}
\end{center}
\caption{A 2-dimensional PCA of the representations produced by the reservoir in Figure~\ref{fig:rsm}
for stacks up to length 10. The color and shape represents the symbol on top of the stack.
Lines indicate the linear separability of representations according to the top symbol.}
\label{fig:states}
\end{figure}

As an example, let us construct an RSM that simulates the LR(1)-automaton $\aut_{a^nb^n}$
in Equation~\ref{eq:anbn}. For such a simulation, we only need to distinguish three cases:
First, if the top of the stack is $b$, we need to pop $3$ symbols from the stack and push the symbol $S$;
second, if the top of the stack is $a$ and the current input symbol is $b$, we need to pop $0$ symbols
from the stack and push the symbol $S$; third, in all other cases we neither pop nor push.
Distinguishing these cases is easy with an echo state network. For example, consider a cycle reservoir
with jumps (CRJ) \cite{CRJ} with five neurons, jump length 2, input weight $1$ and cycle weight $0.5$,
i.e.\ the architecture displayed in Figure~\ref{fig:rsm}.
Figure~\ref{fig:states} shows a two-dimensional principal component analysis
of the states generated by this CRJ for one-hot-coding sequences of length up to 10, where color and
shape indicate the last symbol of the sequence. As we can see, the states are
linearly separable based on their last symbol. Accordingly, we can implement $c^\text{pop}$
and $c^\text{push}$ as linear classifiers, $c^\text{shift}$ as a constant $1$, and $c^\text{out}$
merely needs to recognize whether the current stack content is exactly $S$ or anything else.
Then, our reservoir stack machine will return $1$ for any word that is in $\lng_{a^nb^n}$
and $0$ otherwise.

This example illustrates how the reservoir stack machine is intended to work: The input state $\vec h_t$
represents the current input symbol and the stack state $\vec g_t$ a suffix of the stack, such that
simple classifiers can make the decision which rule of an LR(1)-automaton to apply.
Next, we generalize this idea to a theorem, showing how to simulate an LR(1)-automaton with an RSM
more generally.

\subsection{Theory}
\label{sec:theory}

This section has two goals: First, we wish to show that an echo state network with a constant-sized
memory as proposed in \citet{RMM2} is not sufficient to simulate general LR(1) automata. Second, we
wish to show that the reservoir stack machine is sufficient, provided that the underlying reservoir
is sufficiently rich.

We begin with the first claim. First, we recall the definition of a reservoir memory machine from
\citet{RMM2} (in a slightly adapted form for our notation in this paper).

\begin{dfn}[Reservoir Memory Machine \cite{RMM2}]
Let $\Sigma \subset \R^n$ be a compact set. We define a reservoir memory machine (RMM) over $\Sigma$ with $m$ neurons,
$K$ rows of memory and $L$ outputs as a $5$-tuple of the form
$\mathcal{M} = (\bm{U}, \bm{W}, \sigma, c, g)$, where $(\bm{U}, \bm{W}, \sigma)$ is a reservoir with
$n$ inputs and $m$ neurons that fulfills the echo state property with respect to $\Sigma$, where
$c : \R^m \to \{0, \ldots, K\}$, and where $g : \R^m \to \R^L$.

Let $\vec x_1, \ldots, \vec x_T \in \Sigma^*$. We define the output sequence
$\mathcal{M}(\vec x_1, \ldots, \vec x_T) = \vec y_1, \ldots, \vec x_T$ as the output of
Algorithm~\ref{alg:rmm} for the input $\vec x_1, \ldots, \vec x_T$..
\end{dfn}

\begin{algorithm}
\caption{The dynamics of a reservoir memory machine $\mathcal{M} = (\bm{U}, \bm{W}, \sigma, c, g)$
on the input sequence $\vec x_1, \ldots, \vec x_T$.}
\label{alg:rmm}
\begin{algorithmic}[1]
\Function{RMM}{Input sequence $\vec x_1, \ldots, \vec x_T$}
\State Initialize an empty memory matrix $\bm{M}$ of size $K \times m$ with zeros.
\State Initialize $\vec h_0$.
\For{$t \gets 1, \ldots, T$}
\State $\vec h_t \gets \sigma\big( \bm{W} \cdot \vec h_{t-1} + \bm{U} \cdot \vec x_t\big)$.
\State $a_t \gets c(\vec h_t)$.
\If{$a_t > 0$}
\If{$\vec m_{a_t}$ has already been written to}
\State $\vec h_t \gets \vec m_{a_t}$.
\Else
\State $\vec m_{a_t} \gets \vec h_t$.
\EndIf
\EndIf
\State $\vec y_t \gets g(\vec h_t)$.
\EndFor
\State \Return $\vec y_1, \ldots, \vec y_T$.
\EndFunction
\end{algorithmic}
\end{algorithm}

Intuitively, the classifier $c$ controls which memory location we are currently accessing and the
function $g$ controls the output of the system. As long as $c$ outputs zero, the RMM behaves like a
standard ESN. When it outputs a nonzero memory address $a_t$ the first time at time $t$, we write the
current state $\vec h_t$ to the $a_t$th row of the memory. When the same address $a_{t'}$ is accessed again
at time $t' > t$, we override $\vec h_{t'}$ with the memory entry at $a_{t'}$. In other words, the output
will be the same as that of an ESN until we have two times $t$, $t'$ with $t' > t$
such that $a_t = a_{t'} > 0$. As soon as that happens, the RMM recalls past states.

This memory mechanism is provably sufficient to simulate any finite state automaton \cite{RMM2}.
However, it is insufficient to simulate at least some LR(1) automata. In particular, 
consider the following LR(1) automaton:
\begin{equation}
\aut_\text{palin} = \Big( \{S\}, \{a, b, \$\}, [(\$, \varepsilon, 1, S), (aSa, \varepsilon, 3, S), (bSb, \varepsilon, 3, S)], \{S\} \Big). \label{eq:palindromes}
\end{equation}
This automaton recognizes the language $\lng(\aut_\text{palin})$ of palindromes over $a$ and $b$
with $\$$ in the middle. We will now show that no RMM can simulate this automaton.

\begin{thm}\label{thm:rmm}
Let $\Sigma = \{a, b, \$\}$ be a set of $n$-dimensional vector encodings of the symbols $a$, $b$
and $\$$. Further, let $\mathcal{M} = (\bm{U}, \bm{W}, \sigma, c, g)$ be a reservoir memory
machine with $m$ neurons, $K$ rows of memory, and $L = 1$ outputs where $g$ is a uniformly
continuous output function, i.e.\ for any $\epsilon > 0$ there exists some $\tilde \delta_\epsilon$
such that for any two $\vec h, \vec h' \in \R^m$ with $\lVert \vec h - \vec h'\rVert < \tilde \delta_\epsilon$
it holds $|g(\vec h) - g(\vec h')| < \epsilon$. Then, $\mathcal{M}$ does not simulate $\aut_\text{palin}$.

\begin{proof}
We perform a proof by contradiction, i.e.\ we assume that there exists an RMM $\mathcal{M}$ with a
uniformly continuous output function $g$ and which does simulate $\aut_\text{palin}$ and we
show that this yields a contradiction.

Our proof has two steps. First, we show that any RMM that simulates $\aut_\text{palin}$ never
visits line 9 of Algorithm~\ref{alg:rmm} for inputs of the form $a^T \$ a^T$ or $b a^{T-1} \$ a^{T-1} b$,
i.e.\ it behaves like an ESN for these inputs. Then, we show that a reservoir with the echo state
property can not distinguish the inputs $a^T \$ a^T$ and $ba^T \$ a^T$ for large enough $T$, which in
turn means that an ESN with uniformly continuous output function can not distinguish them, either.

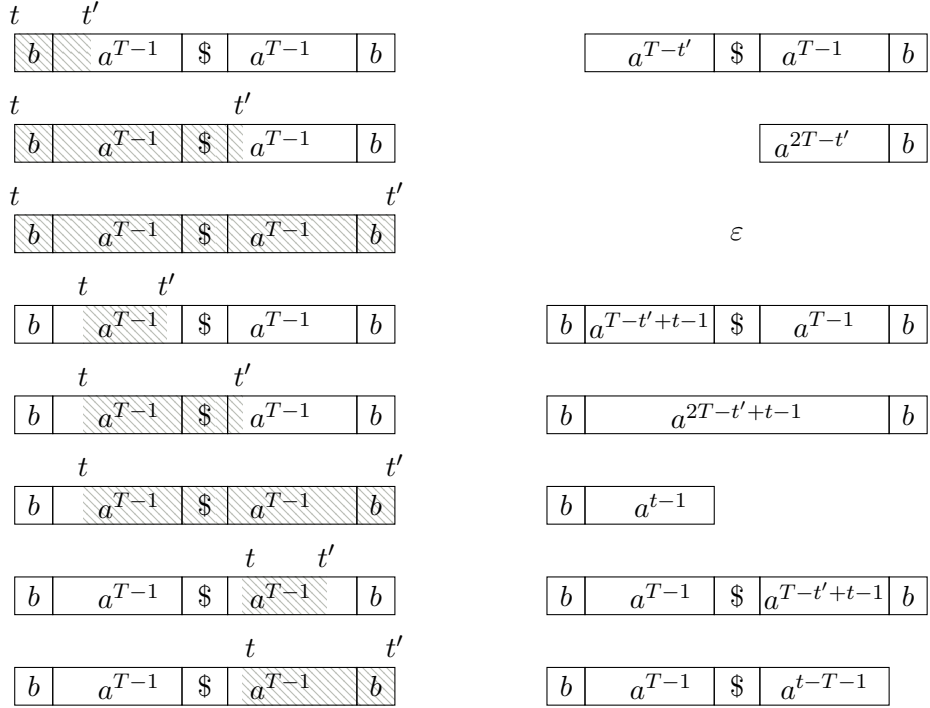
\begin{figure}
\begin{center}
\begin{tikzpicture}
\node at (0,0) {Construction for $a^T\$a^T$};

\begin{scope}[shift={(-3.5,-2)}]


\node at (0,+1) {$\aut_\text{palin}(\bar x) = 1$};

\draw[pattern=north west lines, pattern color=aluminium3, draw=none] (-1.4,+0.25) rectangle (-0.7,-0.25);

\draw (-2.5,+0.25) rectangle (-0.3,-0.25);
\draw (-0.3,+0.25) rectangle (+0.3,-0.25);
\draw (+0.3,+0.25) rectangle (+2.5,-0.25);

\node at (-1,0) {$a^T$};
\node at (0,0) {$\$$};
\node at (+1,0) {$a^T$};

\node[above] at (-1.4,+0.25) {$t$};
\node[above] at (-0.7,+0.25) {$t'$};

\end{scope}

\begin{scope}[shift={(+3.5,-2)}]

\node at (0,+1) {$\aut_\text{palin}(\bar x') = 0$};

\draw (-1.8,+0.25) rectangle (-0.3,-0.25);
\draw (-0.3,+0.25) rectangle (+0.3,-0.25);
\draw (+0.3,+0.25) rectangle (+2.5,-0.25);

\node at (-1,0) {$a^{T-t'+t}$};
\node at (0,0) {$\$$};
\node at (+1,0) {$a^T$};

\end{scope}

\begin{scope}[shift={(-3.5,-3.2)}]


\draw[pattern=north west lines, pattern color=aluminium3, draw=none] (-0.7,+0.25) rectangle (+0.4,-0.25);

\draw (-2.5,+0.25) rectangle (-0.3,-0.25);
\draw (-0.3,+0.25) rectangle (+0.3,-0.25);
\draw (+0.3,+0.25) rectangle (+2.5,-0.25);

\node at (-1,0) {$a^T$};
\node at (0,0) {$\$$};
\node at (+1,0) {$a^T$};

\node[above] at (-0.7,+0.25) {$t$};
\node[above] at (+0.4,+0.25) {$t'$};

\end{scope}

\begin{scope}[shift={(+3.5,-3.2)}]

\draw (-1.8,+0.25) rectangle (+2.1,-0.25);

\node at (0,0) {$a^{2T - t' + t + 1}$};


\end{scope}

\begin{scope}[shift={(-3.5,-4.4)}]


\draw[pattern=north west lines, pattern color=aluminium3, draw=none] (+0.7,+0.25) rectangle (+1.5,-0.25);

\draw (-2.5,+0.25) rectangle (-0.3,-0.25);
\draw (-0.3,+0.25) rectangle (+0.3,-0.25);
\draw (+0.3,+0.25) rectangle (+2.5,-0.25);

\node at (-1,0) {$a^T$};
\node at (0,0) {$\$$};
\node at (+1,0) {$a^T$};

\node[above] at (+0.7,+0.25) {$t$};
\node[above] at (+1.5,+0.25) {$t'$};

\end{scope}

\begin{scope}[shift={(+4.2,-4.4)}]

\draw (-2.5,+0.25) rectangle (-0.3,-0.25);
\draw (-0.3,+0.25) rectangle (+0.3,-0.25);
\draw (+0.3,+0.25) rectangle (+2.0,-0.25);

\node at (-1,0) {$a^T$};
\node at (0,0) {$\$$};
\node at (+1,0) {$a^{T-t'+t}$};

\end{scope}

\node at (0,-6) {Construction for $ba^{T-1}\$a^{T-1}b$};

\begin{scope}[shift={(-3.5,-7)}]


\draw[pattern=north west lines, pattern color=aluminium3, draw=none] (-2.5,+0.25) rectangle (-1.5,-0.25);

\draw (-2.5,+0.25) rectangle (-2.0,-0.25);
\draw (-2.0,+0.25) rectangle (-0.3,-0.25);
\draw (-0.3,+0.25) rectangle (+0.3,-0.25);
\draw (+0.3,+0.25) rectangle (+2.0,-0.25);
\draw (+2.0,+0.25) rectangle (+2.5,-0.25);

\node at (-2.25,0) {$b$};
\node at (-1,0) {$a^{T-1}$};
\node at (0,0) {$\$$};
\node at (+1,0) {$a^{T-1}$};
\node at (+2.25,0) {$b$};

\node[above] at (-2.5,+0.25) {$t$};
\node[above] at (-1.5,+0.25) {$t'$};

\end{scope}

\begin{scope}[shift={(+3.5,-7)}]

\draw (-2.0,+0.25) rectangle (-0.3,-0.25);
\draw (-0.3,+0.25) rectangle (+0.3,-0.25);
\draw (+0.3,+0.25) rectangle (+2.0,-0.25);
\draw (+2.0,+0.25) rectangle (+2.5,-0.25);

\node at (-1,0) {$a^{T-t'}$};
\node at (0,0) {$\$$};
\node at (+1,0) {$a^{T-1}$};
\node at (+2.25,0) {$b$};

\end{scope}

\begin{scope}[shift={(-3.5,-8.2)}]


\draw[pattern=north west lines, pattern color=aluminium3, draw=none] (-2.5,+0.25) rectangle (+0.5,-0.25);

\draw (-2.5,+0.25) rectangle (-2.0,-0.25);
\draw (-2.0,+0.25) rectangle (-0.3,-0.25);
\draw (-0.3,+0.25) rectangle (+0.3,-0.25);
\draw (+0.3,+0.25) rectangle (+2.0,-0.25);
\draw (+2.0,+0.25) rectangle (+2.5,-0.25);

\node at (-2.25,0) {$b$};
\node at (-1,0) {$a^{T-1}$};
\node at (0,0) {$\$$};
\node at (+1,0) {$a^{T-1}$};
\node at (+2.25,0) {$b$};

\node[above] at (-2.5,+0.25) {$t$};
\node[above] at (+0.5,+0.25) {$t'$};

\end{scope}

\begin{scope}[shift={(+3.5,-8.2)}]

\draw (+0.3,+0.25) rectangle (+2.0,-0.25);
\draw (+2.0,+0.25) rectangle (+2.5,-0.25);

\node at (+1,0) {$a^{2T-t'}$};
\node at (+2.25,0) {$b$};

\end{scope}

\begin{scope}[shift={(-3.5,-9.4)}]


\draw[pattern=north west lines, pattern color=aluminium3, draw=none] (-2.5,+0.25) rectangle (+2.5,-0.25);

\draw (-2.5,+0.25) rectangle (-2.0,-0.25);
\draw (-2.0,+0.25) rectangle (-0.3,-0.25);
\draw (-0.3,+0.25) rectangle (+0.3,-0.25);
\draw (+0.3,+0.25) rectangle (+2.0,-0.25);
\draw (+2.0,+0.25) rectangle (+2.5,-0.25);

\node at (-2.25,0) {$b$};
\node at (-1,0) {$a^{T-1}$};
\node at (0,0) {$\$$};
\node at (+1,0) {$a^{T-1}$};
\node at (+2.25,0) {$b$};

\node[above] at (-2.5,+0.25) {$t$};
\node[above] at (+2.5,+0.25) {$t'$};

\end{scope}

\begin{scope}[shift={(+3.5,-9.4)}]

\node at (0,0) {$\varepsilon$};

\end{scope}

\begin{scope}[shift={(-3.5,-10.6)}]


\draw[pattern=north west lines, pattern color=aluminium3, draw=none] (-1.6,+0.25) rectangle (-0.5,-0.25);

\draw (-2.5,+0.25) rectangle (-2.0,-0.25);
\draw (-2.0,+0.25) rectangle (-0.3,-0.25);
\draw (-0.3,+0.25) rectangle (+0.3,-0.25);
\draw (+0.3,+0.25) rectangle (+2.0,-0.25);
\draw (+2.0,+0.25) rectangle (+2.5,-0.25);

\node at (-2.25,0) {$b$};
\node at (-1,0) {$a^{T-1}$};
\node at (0,0) {$\$$};
\node at (+1,0) {$a^{T-1}$};
\node at (+2.25,0) {$b$};

\node[above] at (-1.6,+0.25) {$t$};
\node[above] at (-0.5,+0.25) {$t'$};

\end{scope}

\begin{scope}[shift={(+3.5,-10.6)}]

\draw (-2.5,+0.25) rectangle (-2,-0.25);
\draw (-2,+0.25) rectangle (-0.3,-0.25);
\draw (-0.3,+0.25) rectangle (+0.3,-0.25);
\draw (+0.3,+0.25) rectangle (+2,-0.25);
\draw (+2,+0.25) rectangle (+2.5,-0.25);

\node at (-2.25,0) {$b$};
\node at (-1.15,0) {$a^{T-t'+t-1}$};
\node at (0,0) {$\$$};
\node at (+1.15,0) {$a^{T-1}$};
\node at (+2.25,0) {$b$};

\end{scope}

\begin{scope}[shift={(-3.5,-11.8)}]


\draw[pattern=north west lines, pattern color=aluminium3, draw=none] (-1.6,+0.25) rectangle (+0.5,-0.25);

\draw (-2.5,+0.25) rectangle (-2.0,-0.25);
\draw (-2.0,+0.25) rectangle (-0.3,-0.25);
\draw (-0.3,+0.25) rectangle (+0.3,-0.25);
\draw (+0.3,+0.25) rectangle (+2.0,-0.25);
\draw (+2.0,+0.25) rectangle (+2.5,-0.25);

\node at (-2.25,0) {$b$};
\node at (-1,0) {$a^{T-1}$};
\node at (0,0) {$\$$};
\node at (+1,0) {$a^{T-1}$};
\node at (+2.25,0) {$b$};

\node[above] at (-1.6,+0.25) {$t$};
\node[above] at (+0.5,+0.25) {$t'$};

\end{scope}

\begin{scope}[shift={(+3.5,-11.8)}]

\draw (-2.5,+0.25) rectangle (-2,-0.25);
\draw (-2,+0.25) rectangle (+2.0,-0.25);
\draw (+2,+0.25) rectangle (+2.5,-0.25);

\node at (-2.25,0) {$b$};
\node at (0,0) {$a^{2T-t'+t-1}$};
\node at (+2.25,0) {$b$};

\end{scope}

\begin{scope}[shift={(-3.5,-13)}]


\draw[pattern=north west lines, pattern color=aluminium3, draw=none] (-1.6,+0.25) rectangle (+2.5,-0.25);

\draw (-2.5,+0.25) rectangle (-2.0,-0.25);
\draw (-2.0,+0.25) rectangle (-0.3,-0.25);
\draw (-0.3,+0.25) rectangle (+0.3,-0.25);
\draw (+0.3,+0.25) rectangle (+2.0,-0.25);
\draw (+2.0,+0.25) rectangle (+2.5,-0.25);

\node at (-2.25,0) {$b$};
\node at (-1,0) {$a^{T-1}$};
\node at (0,0) {$\$$};
\node at (+1,0) {$a^{T-1}$};
\node at (+2.25,0) {$b$};

\node[above] at (-1.6,+0.25) {$t$};
\node[above] at (+2.5,+0.25) {$t'$};

\end{scope}

\begin{scope}[shift={(+3.5,-13)}]

\draw (-2.5,+0.25) rectangle (-2,-0.25);
\draw (-2.0,+0.25) rectangle (-0.3,-0.25);

\node at (-2.25,0) {$b$};
\node at (-1,0) {$a^{t-1}$};

\end{scope}

\begin{scope}[shift={(-3.5,-14.2)}]


\draw[pattern=north west lines, pattern color=aluminium3, draw=none] (+0.5,+0.25) rectangle (+1.6,-0.25);

\draw (-2.5,+0.25) rectangle (-2.0,-0.25);
\draw (-2.0,+0.25) rectangle (-0.3,-0.25);
\draw (-0.3,+0.25) rectangle (+0.3,-0.25);
\draw (+0.3,+0.25) rectangle (+2.0,-0.25);
\draw (+2.0,+0.25) rectangle (+2.5,-0.25);

\node at (-2.25,0) {$b$};
\node at (-1,0) {$a^{T-1}$};
\node at (0,0) {$\$$};
\node at (+1,0) {$a^{T-1}$};
\node at (+2.25,0) {$b$};

\node[above] at (+0.6,+0.25) {$t$};
\node[above] at (+1.6,+0.25) {$t'$};

\end{scope}

\begin{scope}[shift={(+3.5,-14.2)}]

\draw (-2.5,+0.25) rectangle (-2.0,-0.25);
\draw (-2.0,+0.25) rectangle (-0.3,-0.25);
\draw (-0.3,+0.25) rectangle (+0.3,-0.25);
\draw (+0.3,+0.25) rectangle (+2.0,-0.25);
\draw (+2.0,+0.25) rectangle (+2.5,-0.25);

\node at (-2.25,0) {$b$};
\node at (-1,0) {$a^{T-1}$};
\node at (0,0) {$\$$};
\node at (+1.15,0) {$a^{T-t'+t-1}$};
\node at (+2.25,0) {$b$};

\end{scope}

\begin{scope}[shift={(-3.5,-15.4)}]


\draw[pattern=north west lines, pattern color=aluminium3, draw=none] (+0.5,+0.25) rectangle (+2.5,-0.25);

\draw (-2.5,+0.25) rectangle (-2.0,-0.25);
\draw (-2.0,+0.25) rectangle (-0.3,-0.25);
\draw (-0.3,+0.25) rectangle (+0.3,-0.25);
\draw (+0.3,+0.25) rectangle (+2.0,-0.25);
\draw (+2.0,+0.25) rectangle (+2.5,-0.25);

\node at (-2.25,0) {$b$};
\node at (-1,0) {$a^{T-1}$};
\node at (0,0) {$\$$};
\node at (+1,0) {$a^{T-1}$};
\node at (+2.25,0) {$b$};

\node[above] at (+0.6,+0.25) {$t$};
\node[above] at (+2.5,+0.25) {$t'$};

\end{scope}

\begin{scope}[shift={(+3.5,-15.4)}]

\draw (-2.5,+0.25) rectangle (-2.0,-0.25);
\draw (-2.0,+0.25) rectangle (-0.3,-0.25);
\draw (-0.3,+0.25) rectangle (+0.3,-0.25);
\draw (+0.3,+0.25) rectangle (+2.0,-0.25);

\node at (-2.25,0) {$b$};
\node at (-1,0) {$a^{T-1}$};
\node at (0,0) {$\$$};
\node at (+1.15,0) {$a^{t-T-1}$};

\end{scope}

\end{tikzpicture}
\end{center}
\caption{All possibilities to recall memory in an RMM at different locations of processing inputs of the form
$a^T\$a^T$ (top) or $ba^{T-1}\$a^{T-1}b$ (bottom). In all inputs on the left, the shaded region indicates the portion
of the input that is ignored and the input on the right corresponds to a version of the input where
the shaded region is removed. In all cases, we observe that the input on the left would be accepted
by $\aut_\text{palin}$, while the input on the right would not. This is a contradiction, because
both inputs would receive the same state (and hence the same output) by the RMM.}
\label{fig:rmm_proof}
\end{figure}

Regarding our first claim, assume that line 9 of of Algorithm~\ref{alg:rmm} is visited.
Then, there must exist $t, t'$ such that $0 < t < t' \leq 2T + 1$ and $a_t = a_{t'} > 0$.
Now, let $t'$ be as large as possible and $t$ be as small as possible.
In that case, the original sequence $\bar x$ and $\bar x'$ with the subsequence from $t$ to $t'$
removed yield the same state (refer to \citet{RMM2}) and, accordingly, the same output. However,
we can show that $\bar x$ would be accepted by $\aut_\text{palin}$
while $\bar x'$ would not be accepted. Accordingly, the RMM does not simulate $\aut_\text{palin}$,
which is a contradiction. Refer to Figure~\ref{fig:rmm_proof} for an illustration of all possible
choices of $t$ and $t'$.

By extension of this argument, we also know that line 9 is not visited for the prefix
$ba^T\$a^T$, otherwise it would also be visited for $ba^T\$a^Tb$. Further,
given that line 9 is not visited, the output generated by Algorithm~\ref{alg:rmm} is the same as
the output of the ESN $(\bm{U}, \bm{W}, \sigma, g)$ according to Definition~\ref{dfn:esn},
because lines 6-8 and 10-13 neither influence the state $\vec h_t$ nor the output.

Next, we show that the ESN $(\bm{U}, \bm{W}, \sigma, g)$ can not simulate the automaton
$\aut_\text{palin}$. For this purpose,
we consider the output function $g$ and the echo state property in a bit more detail. Because
$g$ is uniformly continuous, we know that there must exist some $\tilde \delta$,
such that for any two $\vec h, \vec h' \in \R^m$ with $\lVert \vec h - \vec h' \rVert < \tilde \delta$
we obtain $|g(\vec h) - g(\vec h')| < 1$. Next, let $\delta_1, \delta_2, \ldots$ be the null sequence
from the definition of the echo state property (refer to Definition~\ref{dfn:esn}). Because this is a
null sequence, there must exist some $t^*$ such that for any $t > t^*$ we obtain
$\delta_t < \tilde \delta$. Now, consider the two inputs $a^T\$a^T$ and $ba^T\$a^T$ with
$T = \lceil t^* / 2 \rceil$ and consider the two initial states $\vec h_0$ and
$\vec h'_0 = \sigma(\bm{U} \cdot b)$, as well as the continuations
$\vec h_t = \sigma(\bm{W} \cdot \vec h_{t-1} + \bm{U} \cdot \vec x_t)$ and
$\vec h'_t = \sigma(\bm{W} \cdot \vec h'_{t-1} + \bm{U} \cdot \vec x_t)$
for all $t \in \{1, \ldots, 2T+1\}$ where $\vec x_1, \ldots, \vec x_{2T+1} = a^T\$a^T$.

The echo state property now guarantees that $\lVert \vec h_{2T+1} - \vec h'_{2T+1} \rVert \leq \delta_{2T+1}
< \tilde \delta$ because $2T+1 > t^*$. Accordingly, the uniform continuity of $g$ guarantees that
$|g(\vec h_{2T+1}) - g(\vec h'_{2T+1})| < 1$. However, we have
$|\aut_\text{palin}(ba^T\$a^T) - \aut_\text{palin}(a^T\$a^T)| = 1$.
Accordingly, $\mathcal{M}$ does not simulate $\aut_\text{palin}$.
\end{proof}
\end{thm}

Our next goal is to prove that an RSM with a sufficiently rich reservoir can simulate any LR(1)-automaton,
only by adjusting the functions $c^\text{pop}$, $c^\text{push}$, and
$c^\text{out}$ and $c^\text{shift}$. To make this precise,
we first define formally what we mean by a 'sufficiently rich reservoir' and then go on to our main theorem.

\begin{dfn}[$\bar w$-separating reservoirs]
Let $\Sigma \subset \R^n$ be some finite set and $\res = (\bm{U}, \bm{W}, \sigma)$ be a
reservoir with $n$ inputs and $m$ neurons. We define the representation
$h_\res(\vec x_1, \ldots, \vec x_T)$ of $\vec x_1, \ldots, \vec x_T \in \Sigma^*$
according to the reservoir $\res$ as the state $\vec h_T$ resulting from Equation~\ref{eq:rnn}.

Now, let $\bar w \in \Sigma^*$. We say that $\res$ is a
\emph{$\bar w$-separating reservoir} if there exists an affine function
$f_{\bar w}(\vec h) = \bm{V} \cdot \vec h + b$ such that for all $\bar u \in \Sigma^*$
it holds: $f_{\bar w}(h_\res(\bar u)) > 0$ if $\bar u$ has $\bar w$ as suffix and
$f_{\bar w}(h_\res(\bar u)) \leq 0$ otherwise.
\end{dfn}

For example, Figure~\ref{fig:states} shows the representations of random words
$\bar w \in \{a, b, S\}^*$ up to length 10.
The figure shows that the reservoir is $a$-separating, $b$-separating,
and $S$-separating. In general, reservoirs with contractive properties have
a strong bias to be separating, because the suffix dominates the representation
\cite{FractalESN}. We also use this separation property to simulate LR(1)-automata with RSMs.

\begin{thm}\label{thm:lr_rsm}
Let $\aut = (\Phi, \Sigma, R, F)$ be an LR(1)-automaton with $\Phi \subset \R^n$ and
$\Sigma \subset \R^n$. Futher, let
$\mathcal{X} = \{ x | (\bar s, x, j, A) \in R\}$, and let
$\mathcal{S} = \{ \bar s | (\bar s, x, j, A) \in R\}$.
Finally, let $\res = (\bm{U}, \bm{W}, \sigma)$ be a reservoir over $\Sigma \cup \Phi$ with
$n$ inputs and $m$ neurons such that $\res$ is a
$\bar w$-separator for all $\bar w \in \mathcal{X} \cup \mathcal{S}$ and such that
$\{ h_\res(\bar w) | \bar w \in (\Phi \cup\Sigma)^*, \bar w \notin F \} \cap
\{ h_\res(A) | A \in F\} = \emptyset$, i.e.\ the representations of accepting nonterminals
and all other words are disjoint.

Then, we can construct classifiers $c^\text{pop}$, $c^\text{push}$, $c^\text{shift}$,
and $c^\text{out}$, such that the reservoir stack machine
$(\bm{U}, \bm{W}, \sigma, c^\text{pop}, c^\text{push}, c^\text{shift}, c^\text{out})$
over $\Sigma$ and  $\Phi$ simulates $\aut$.

\begin{proof}
Per definition, $\res$ is a $\bar w$-separator for every word
$\bar w \in \mathcal{X} \cup \mathcal{S}$.
Accordingly, there must exist affine functions $f_{\bar w} : \R^m \to \R$,
such that for all $\bar u \in \Sigma^*$ it holds: $f_{\bar w}(h_\res(\bar u)) > 1$ if $\bar w$ is a
suffix of $\bar u$ and $f_{\bar w}(h_\res(\bar u)) \leq 0$ otherwise.

Now, we define the functions $c^\text{pop}$ and $c^\text{push}$ via the following procedure.
We iterate over the rules $(\bar s, x, j, A) \in R$ of the LR(1)-automaton
in ascending order. If $f_{\bar s}(\vec g_t) > 0$ and $f_x(\vec h_t) > 0$, we return
$c^\text{pop}(\vec h_t, \vec g_t) = j$ and $c^\text{push}(\vec h_t, \vec g_t) = A$.
If this does not occur for any rule, we return
$c^\text{pop}(\vec h_t, \vec g_t)= 0$ and $c^\text{push}(\vec h_t, \vec g_t) = \vec 0$.

We further define $c^\text{shift}(\vec h_t, \vec g_t)$ as constant $1$.
Finally, we define $c^\text{out}(\vec h_t, \vec g_t)$ as $1$ if
$\vec g_t \in \{ h_\res(A) | A \in F\}$ and as $0$ otherwise. Note that all four functions
are well-defined due to our separation requirements on the reservoir.

Our claim is now that the stack $\mathcal{S}$ when arriving at line 25 in Algorithm~\ref{alg:rsm}
is the same as the stack $\mathcal{S}$ when arriving at line 8 in Algorithm~\ref{alg:parse}.
We prove this by induction over the number of inner loop iterations.
If no iterations have occurred, the stack is empty in both cases. Now, assume that the stack
has a certain state $\mathcal{S}$ in both Algorithm~\ref{alg:parse} and Algorithm~\ref{alg:rsm}
and we now enter line 4 in Algorithm~\ref{alg:parse} and line 8 in Algorithm~\ref{alg:rsm},
respectively.

First, consider the case that the $\exists$ quantifier in line 4 of Algorithm~\ref{alg:parse}
finds no matching rule, i.e.\ there is no rule $(\bar s, x, j, A) \in R$ such that the current
stack $\mathcal{S}$ has the suffix $\bar s$ and the current input symbol $y$ equals $x$ (or $x = \varepsilon$).
In that case, $f_{\bar s}(\vec g_t) \leq 0$ or $f_x(\vec h_t) \leq 0$, which implies
$c^\text{pop}(\vec h_t, \vec g_t) = 0$ and $c^\text{push}(\vec h_t, \vec g_t) = \vec 0$. Accordingly, 
lines 9-20 leave the stack unchanged and the condition in line 21 applies, such that we break
out of the loop. Equivalently, the loop in Algorithm~\ref{alg:parse} stops.

Second, consider the case that at least one rule matches in line 4 of Algorithm~\ref{alg:parse}.
In that case, the first matching rule $(\bar s, x, j, A)$ is selected (by definition in the
Algorithm description). Accordingly, the current stack $\mathcal{S}$ has the suffix $\bar s$
and the current input symbol $y$ equals $x$ or $x = \varepsilon$. By virtue of the separation
conditions on our reservoir, $f_{\bar s}(\vec g_t) > 0$ and $f_x(\vec h_t) > 0$. Therefore,
by our definition of $c^\text{pop}$ and $c^\text{push}$ above, we obtain
$c^\text{pop}(\vec h_t, \vec g_t) = j$ and $c^\text{push}(\vec h_t, \vec g_t) = A$.

Next, note that lines 10 and 18 of Algorithm~\ref{alg:rsm} update the stack just as lines 5-6 in
Algorithm~\ref{alg:parse} do and that lines 11-14 as well as line 19 update the stack state $\vec g_t$
accordingly.

Finally, note that the condition in line 26 of Algorithm~\ref{alg:rsm} is always fulfilled because we defined
$c^\text{shift}(\vec h_t, \vec g_t) = 1$. Hence, line 27 of Algorithm~\ref{alg:rsm} updates the stack
just as line 8 in Algorithm~\ref{alg:parse} does and line 25 updates the stack representation
accordingly.

In conclusion, at every iteration of the computation, the stacks of Algorithms~\ref{alg:parse} and~\ref{alg:rsm}
are the same. It only remains to show that the last output $\vec y_{T+1}$ is the same as
$\aut(\bar w)$. If $\aut(\bar w) = 1$, this means that the stack at the end of the computation
in Algorithm~\ref{alg:parse} was $\mathcal{S} = A\#$ for some accepting nonterminal $A \in F$.
For Algorithm~\ref{alg:rsm}, this implies that the stack before the last push operation
in line 24 must have been $\mathcal{S} = A$. Hence,
$c^\text{out}(\vec h_{T+1}, \vec g_{T+1}) = 1$. Conversely, if $\aut(\bar w) = 0$, the stack must have
been something else, and hence $c^\text{out}(\vec h_{T+1}, \vec g_{T+1}) = 0$.
\end{proof}
\end{thm}

We note that we assume linear separability of single suffices, but we may still require nonlinear
classifiers for $c^\text{pop}$, $c^\text{push}$, and $c^\text{out}$ because these are defined over unions of suffices,
which in turn may yield non-linear boundaries. As such, we recommend non-linear classifiers in practice,
such as radial basis function SVMs.

This concludes our theory chapter. We now know that RMMs cannot recognize LR(1) languages in general,
but RSMs can. Our next step is to describe how we can train an RSM.

\subsection{Training}
\label{sec:training}

\begin{algorithm}
\caption{An algorithm to re-order training data from a $5$-tuple
$(\bar x, \bm{J}, \bm{A}, \vec \rho, \bm{Y})$ of inputs $\bar x \in \Sigma^T$,
desired pop actions $\bm{J} \in \{0, \ldots, J\}^{T+1 \times M}$,
desired push actions $\bm{A} \in \Phi^{T+1 \times M}$,
desired shift actions $\vec \rho \in \{0, 1\}^{T+1}$, and
desired outputs $\bm{Y} \in \R^{T+1 \times L}$ 
into input-output pairs for training the functions
$c^\text{pop}$, $c^\text{push}$, $c^\text{shift}$, and $c^\text{out}$.
We assume that a reservoir $(\bm{U}, \bm{W}, \sigma)$
with $\bm{U} \in \R^{m \times (n + K)}$, $\bm{W} \in \R^{m \times m}$, and $\sigma : \R \to \R$.}
\label{alg:training}
\begin{algorithmic}[1]
\Function{RSM-train}{Input training data $(\bar x, \bm{J}, \bm{A}, \vec \rho, \bm{Y})$}
\State Initialize $H^\text{pop/push}$, $Y^\text{pop}$, and $Y^\text{push}$ as empty lists.
\State Initialize an empty stack $\mathcal{S}$. Initialize $\vec h_0 \gets \vec 0$ and $\vec g_1 \gets \vec 0$.
\For{$t \gets \{1, \ldots, T+1\}$}
\State $\vec h_t \gets \sigma\Big(\bm{U} \cdot \vec x_t + \bm{W} \cdot \vec h_{t-1}\Big)$.
\For{$\tau \gets \{1, \ldots, M\}$}
\State Append $(\vec h_t, \vec g_t)$ to $H^\text{pop/push}$.
\State Append $j_{t, \tau}$ to $Y^\text{pop}$ and $\vec a_{t, \tau}$ to $Y^\text{push}$.
\State Pop $j_{t, \tau}$ elements from $\mathcal{S}$.
\If{$\vec a_{t, \tau} \neq \vec 0$}
\State Push $\vec a_{t, \tau}$ onto $\mathcal{S}$.
\EndIf
\State $\vec g_t \gets \vec 0$.
\For{$\vec s \gets $ elements of $\mathcal{S}$}
\State $\vec g_t \gets \sigma\Big(\bm{U} \cdot \vec s + \bm{W} \cdot \vec g_t \Big)$.
\EndFor
\If{$j_{t, \tau} = 0$ and $\vec a_{t, \tau} = \vec 0$}
\State \textbf{Break}.
\EndIf
\EndFor
\If{$\rho_t > 0$}
\State Push $\vec x_t$ onto $\mathcal{S}$.
\State $\vec g_{t+1} \gets \sigma\Big(\bm{U} \cdot \vec x_t + \bm{W} \cdot \vec g_t \Big)$.
\Else
\State $\vec g_{t+1} \gets \vec g_t$.
\EndIf
\EndFor
\State Convert $H^\text{pop/push}$, $Y^\text{pop}$, and $Y^\text{push}$ to matrices.
\State Set up a $T+1 \times 2m$ matrix $\bm{H}$ with rows $(\vec h_t, \vec g_t)$.
\State \Return $(\bm{H}^\text{pop/push}, \bm{Y}^\text{pop})$, $(\bm{H}^\text{pop/push}, \bm{Y}^\text{push})$, $(\bm{H}, \vec \rho)$, $(\bm{H}, \bm{Y})$.
\EndFunction
\end{algorithmic}
\end{algorithm}

In this section, we describe how to train an RSM from example data. As training data, we require
$5$-tuples of the form $(\bar x, \bm{J}, \bm{A}, \vec \rho, \bm{Y})$ where $\bar x \in \Sigma^T$ is a sequence
of $T$ input vectors for some $T$, $\bm{J} \in \{0, \ldots, J\}^{T+1 \times M}$ is a matrix of desired pop actions for some $M$,
$\bm{A} \in \Phi^{T+1 \times M}$ is a tensor of desired push actions,
$\vec \rho \in \{0, 1\}^{T+1}$ is a vector of desired shift actions, and
$\bm{Y} \in \R^{T+1 \times L}$ is a matrix of desired outputs. In other words, our training process
requires ground truth example data for the correct pop, push, and shift behavior in addition to the
desired outputs. However, our model is able to generalize the behavior beyond example data as we will
see later in the experiments.

If we want to simulate the behavior of a known LR(1)-automaton, the training data is simple to generate:
We merely have to execute Algorithm~\ref{alg:parse} and record the rules that are applied, which
directly yield the desired pop and push actions. The desired shift action is constant $1$ and the
desired output is $1$ whenever the stack is $\mathcal{S} = A$ for an accepting nonterminal $A \in F$
and $0$ otherwise. Importantly, we only need to know the behavior of the automaton on short training
examples to learn the general stack interaction from demonstration.

If we do not have an LR(1)-automaton available to generate the training data, we require some other
heuristic, which is dependent on the dataset. In our copy experiments, for example, we use pop and push
actions to normalize the length of the stack at crucial points of the computation, thus making memory
recall much simpler compared to a regular recurrent neural network.

Once training data in form of these tuples is constructed, we apply Algorithm~\ref{alg:training} to
re-order this training data into input-output pairs for each of the functions
$c^\text{pop}$, $c^\text{push}$, $c^\text{shift}$, and $c^\text{out}$. This algorithm is a variation
of Algorithm~\ref{alg:rsm}, controlled via teacher forcing. Finally, once these input-output-pairs
are collected, we can train $c^\text{pop}$, $c^\text{push}$, $c^\text{shift}$, and $c^\text{out}$
via any classification or regression scheme. In this paper, we opt for classic radial basis
function support vector machines (RBF-SVMs)\footnote{The only exception is $c^\text{out}$ in the copy
tasks, where a linear regression is sufficient.} with automatically chosen kernel width as implemented
in scikit-learn \cite{Sklearn}.
Because the objective of RBF-SVMs is convex, the training is fast and globally optimal.

\section{Experiments}
\label{sec:experiments}

We evaluate reservoir stack machines (RSMs) on three benchmark tasks for Neural Turing Machines (NTMs)
\cite{NTM_impl} as well as six context-free languages. In more detail, the three NTM benchmark tasks
are:

\paragraph{latch:} Both input and output are binary and one-dimensional and the
output should jump from zero to one and back whenever there is a one on the input. Equivalently, this
task can be expressed via the regular expression $(0^*10^*1)^*0^*10^*$, i.e.\ we accept any word
that contains an odd number of ones. We sample training words up to length $50$ and test words
starting from length $50$ from a probabilistic context-free grammar%
\footnote{For the precise parameters and experimental details, refer to 
\url{https://gitlab.com/bpaassen/reservoir_stack_machines}.}.

\paragraph{copy:} The input is a 1-20 time step long sequence of 8 random bits, followed by a
special end-of-sequence token on an extra channel (this channel is -1 during the input sequence).
The output should be zero until this extra one, after which the output
should be a copy of the input signal.

\paragraph{repeat copy:} The same as copy but where the input sequence has length up to 10 and
the end-of-sequence token can occur up to 10 times (up to 20 times in the testing data).
After each token, the input sequence should be copied again.

\begin{figure}
\begin{center}
\begin{tikzpicture}
\begin{scope}[shift={(-5,-1.62)}]
\begin{axis}[title={latch task},xlabel={$t$},ylabel={amplitude}, width=5cm, height=4.62cm,
xmin=0, ymax=1.8, legend pos={north east}, legend cell align={left},
ytick={0,1}]
\addplot[class1color, thick] table[x=time,y=x] {latch_example.csv};
\addlegendentry{input}
\addplot[thick, class2color, densely dashed] table[x=time,y=y] {latch_example.csv};
\addlegendentry{output}
\end{axis}
\end{scope}
\begin{groupplot}[view={0}{90}, xlabel={$t$}, ymin=0, ymax=8,
group style={group size=2 by 2,
x descriptions at=edge bottom,y descriptions at=edge left,
horizontal sep=0.7cm, vertical sep=0.2cm},
width=4cm, height=3cm]
\nextgroupplot[title={copy task},ymax=9,width=3.2cm, ylabel={input},
colormap={tango}{color(0cm)=(orange1); color(1cm)=(white); color(2cm)=(skyblue3)}]
\addplot3[surf,mesh/ordering=y varies,mesh/rows=10,shader=flat corner] file {copy_example_input.csv};
\nextgroupplot[title={repeat copy task},ymax=9,
colormap={tango}{color(0cm)=(orange1); color(1cm)=(white); color(2cm)=(skyblue3)}, colorbar]
\addplot3[surf,mesh/ordering=y varies,mesh/rows=10,shader=flat corner] file {repeat_copy_example_input.csv};
\nextgroupplot[width=3.2cm, ylabel={output},
colormap={tango}{color(0cm)=(white); color(1cm)=(skyblue3)}]
\addplot3[surf,mesh/ordering=y varies,mesh/rows=9,shader=flat corner] file {copy_example_output.csv};
\nextgroupplot[colormap={tango}{color(0cm)=(white); color(1cm)=(skyblue3)}]
\addplot3[surf,mesh/ordering=y varies,mesh/rows=9,shader=flat corner] file {repeat_copy_example_output.csv};
\end{groupplot}
\end{tikzpicture}
\vspace{-0.7cm}
\end{center}
\caption{Example input and output sequences for the three NTM benchmark data sets.}
\label{fig:data}
\end{figure}
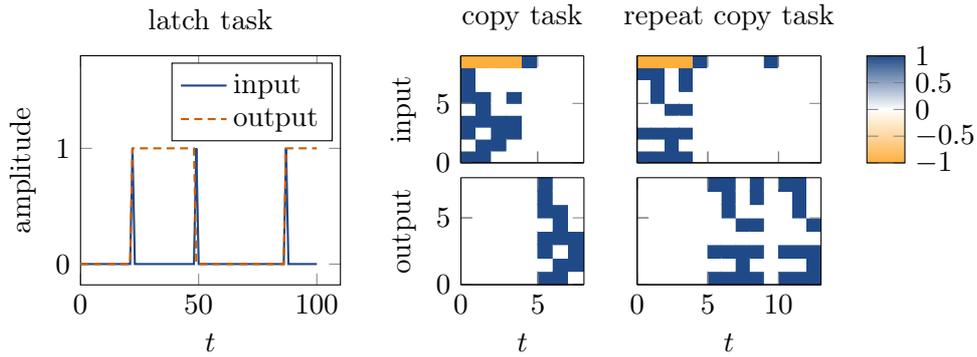

For a graphical illustration of the three datasets, refer to Figure~\ref{fig:data}.
For the latch task, we use the LR(1)-rules $(S0, \varepsilon, 2, S)$, $(S1, \varepsilon, 2, A)$,
$(A0, \varepsilon, 2, A)$, $(A1, \varepsilon, 2, S)$, $(0, \varepsilon, 1, A)$ with the accepting
nonterminal $S$ to generate the desired stack behavior for the training data.
For the copy task, we use the following heuristic: We always shift the input onto the stack,
but we do not use pop or push until the end-of-sequence token appears. At that point, we fill up the
stack with a placeholder nonterminal $S$ until the stack has length $20$. Then, we continue.
By this construction, the output can be constructed by simply copying the $20$th stack element to the
output, i.e.\ the reservoir must be rich enough such that the $20$th stack element can be reconstructed
via $c^\text{out}$ from $\vec g_t$. For this purpose, we use linear ridge regression as implemented in sklearn  
(with the $\alpha$ hyperparameter being part of the hyperparameter optimization)
because a linear operator is provably sufficient to reconstruct past inputs, at least for Legendre
reservoirs \cite{LMU}. 
For repeat copy, we use the same scheme, except for one difference:
If a second end-of-sequence token appears, we pop elements from the stack until no
end-of-sequence token is on the stack anymore and then continue as before.

\begin{table}
\begin{center}
\begin{scriptsize}
\begin{tabular}{llllll}
Dyck1 & Dyck2 & Dyck3 & $a^nb^n$ & Palindromes & JSON \\
\cmidrule(lr){1-3} \cmidrule(lr){4-4} \cmidrule(lr){5-5}\cmidrule(lr){6-6}
$(\nt{SS}, \trm{\varepsilon}, 2, \nt{S})$ & $(\nt{SS}, \trm{\varepsilon}, 2, \nt{S})$ & $(\nt{SS}, \trm{\varepsilon}, 2, \nt{S})$
& $(\trm{a}\nt{S}\trm{b}, \trm{\varepsilon}, 3, \nt{S})$ & $(\trm{a}\nt{S}\trm{a}, \trm{\varepsilon}, 3, \nt{S})$ & $(\trm{\{\}}, \trm{\varepsilon}, 2, \nt{V})$ \\
$(\trm{(}\nt{S}\trm{)}, \trm{\varepsilon}, 3, \nt{S})$ & $(\trm{(}\nt{S}\trm{)}, \trm{\varepsilon}, 3, \nt{S})$ & $(\trm{(}\nt{S}\trm{)}, \trm{\varepsilon}, 3, \nt{S})$ &
$(\trm{a}, \trm{b}, \trm{\varepsilon}, \nt{S})$ & $(\trm{b}\nt{S}\trm{b}, \trm{\varepsilon}, 3, \nt{S})$ & $(\trm{[]}, \trm{\varepsilon}, 2, \nt{V})$  \\
& $(\trm{[}\nt{S}\trm{]}, \trm{\varepsilon}, 3, \nt{S})$ & $(\trm{[}\nt{S}\trm{]}, \trm{\varepsilon}, 3, \nt{S})$ &
& $(\trm{\$}, \trm{\varepsilon}, 1, \nt{S})$ & $(\trm{\{}\nt{O}\trm{\}}, \trm{\varepsilon}, 3, \nt{V})$\\
& & $(\trm{\{}\nt{S}\trm{\}}, \trm{\varepsilon}, 3, \nt{S})$ & & & $(\trm{[}\nt{A}\trm{]}, \trm{\varepsilon}, 3, \nt{V})$ \\
$(\trm{(}, \trm{)}, 0, \nt{S})$ & $(\trm{(}, \trm{)}, 0, \nt{S})$ & $(\trm{(}, \trm{)}, 0, \nt{S})$ & & & $(\trm{n}, \trm{\varepsilon}, 1, \nt{V})$ \\
& $(\trm{[}, \trm{]}, 0, \nt{S})$ & $(\trm{[}, \trm{]}, 0, \nt{S})$ & & & $(\trm{s}, \trm{\varepsilon}, 1, \nt{V})$ \\
& & $(\trm{\{}, \trm{\}}, 0, \nt{S})$ & & & $\trm{(k : }\nt{V}\trm{,} \nt{O}, \trm{\varepsilon}, 5, \nt{O})$ \\
& & & & & $(\trm{k :} \nt{V}, \trm{\}}, 3, \nt{O})$ \\
& & & & & $(\nt{V}\trm{,} \nt{A}, \trm{\varepsilon}, 3, \nt{A})$ \\
& & & & & $(\nt{V}, \trm{]}, 1, \nt{A})$ \\
\cmidrule(lr){1-3} \cmidrule(lr){4-4} \cmidrule(lr){5-5}\cmidrule(lr){6-6}
$\trm{()(())}$ & $\trm{([])[]}$ & $\trm{(\{\})[]}$ & $\trm{aaabbb}$ & $\trm{ab\$ba}$ & $\trm{\{k : [n, n], k : s\}}$ \\
\end{tabular}
\end{scriptsize}
\end{center}
\caption{The list of rules for the LR(1)-automata of all six language data sets and an example word
from each language. Terminal symbols are colored orange, nonterminals blue.}
\label{tab:lrs}
\end{table}

Our six language data sets are:

\paragraph{Dyck1/2/3:} Deterministic context-free languages of balanced bracket pairs with one, two,
and three different kinds of brackets respectively, as suggested by \citet{DeepStacks}.

\paragraph{$a^nb^n$:} The language $\mathcal{L}_{a^nb^n}$ from Equation~\ref{eq:anbn}.

\paragraph{Palindrome:} The language of palindromes over the letters $a$ and $b$ with a $\$$
symbol in the center from Equation~\ref{eq:palindromes}.

\paragraph{JSON:} A simplified version of the javascript object notation (JSON;
\url{https://www.json.org}), where we represent numbers with the symbol $n$, strings with the
symbol $s$, and keys with the symbol $k$. 

The rules of the LR(1)-automata for all six languages are shown in
Table~\ref{tab:lrs}, including an example word from each language. The accepting nonterminals of
the automata are $\{S\}$ for the first five and $\{V\}$ for the last automaton. It is easy to show
that none of these languages is Chomsky-3 by using the pumping lemma \cite{Pumping,TheoInf}.
For each language task, our desired output is $y_t = 1$ if the word up to $t$ is in the language
and $y_t = 0$ otherwise.

We use the ground-truth LR(1)-automaton to annotate the training data with
the desired stack behavior. However, we sample the training words up to length $50$ and
the evaluation words starting from length $50$ as suggested by \citet{DeepStacks},
thus demonstrating that our model can generalize beyond the shown examples.
We sample the words from a probabilistic context-free grammar for the
respective language\footnote{For the precise parameters and experimental details, refer to 
\url{https://gitlab.com/bpaassen/reservoir_stack_machines}.}.

In all experiments, we sample $100$ random sequences for training and another $100$ sequences for
testing. To obtain statistics, we perform $10$ repeats of every experiment.
For hyper-parameter optimization, we sample $20$ random sets of hyperparameters\footnotemark[3], each
evaluated on $3$ separate datasets of $100$ training sequences and $100$ test sequences.

We evaluate three kinds of reservoir for the RSM, namely random Gaussian numbers (rand),
a cycle reservoir with jumps (CRJ) \cite{CRJ}, and a Legendre delay network (LDN) \cite{LMU}. As baseline,
we compare against a standard echo state network \cite{ESN}, which attempts to predict the
desired output via linear regression from the reservoir state $\vec h_t$. We expect that these networks should
fail because they do not have the computational power to recognize languages that require arbitrarily
long memory \cite{DMM}. For all reservoirs, we used $256$
neurons to maintain equal representational power between models\footnote{We observed that $256$ neurons
were insufficient for the copy task. In this case, we increased the number of neurons to $512$.
Further, we report reference results for ESNs with $1024$ neurons in the appendix.}.

We also include three baselines from deep learning, namely a gated recurrent unit (GRU) \cite{GRU},
the stack recurrent neural network (SRNN) \cite{DeepStacks}, and a deep version of our reservoir
stack machine (DSM), where we replace the reservoir with a GRU. For GRUs, we used the implementation
provided by pyTorch \cite{PyTorch} and for SRNNs the reference implementation of Suzgun et
al.\footnote{\url{https://github.com/suzgunmirac/marnns/blob/master/models/rnn_models.py}}.
We trained all deep models with Adam \cite{Adam} with a learning rate of $10^{-3}$, weight
decay of $10^{-8}$, and $10,000$ epochs, where we processed a single word in each epoch.
As for the reservoir models, we used $256$ neurons for each model in a single recurrent
layer.

Note that we do \emph{not} compare against reservoir memory machines (RMMs) \cite{RMM2} because there
is no strategy to generate training data for the LR(1) language datasets (refer to 
Theorem~\ref{thm:rmm}). For the latch, copy, and repeat copy datasets, it has already
been shown that RMMs achieve zero error \cite{RMM2}.

We performed all experiments on a consumer-grade 2017 laptop with Intel i7 CPU.

\begin{table}
\caption{The mean absolute error ($\pm$ std.) on the test data for all models
for the NTM benchmark datasets. All results with mean and standard deviation below $10^{-2}$
are bold-faced.}
\label{tab:benchmark_errors}
\begin{center}
\begin{tabular}{lccc}
model & {latch} & {copy} & {repeat copy} \\
\cmidrule(lr){1-1} \cmidrule(lr){2-4}
rand-ESN & $0.43 \pm 0.02$ & $0.23 \pm 0.00$ & $0.36 \pm 0.01$ \\
CRJ-ESN & $0.40 \pm 0.02$ & $0.20 \pm 0.01$ & $0.34 \pm 0.01$ \\
LDN-ESN & $0.42 \pm 0.01$ & $0.22 \pm 0.00$ & $0.30 \pm 0.01$ \\
\cmidrule(lr){1-1} \cmidrule(lr){2-4}
GRU & $\bm{0.00 \pm 0.00}$ & $0.22 \pm 0.00$ & $0.32 \pm 0.01$ \\
SRNN & $0.30 \pm 0.03$ & $0.21 \pm 0.01$ & $0.36 \pm 0.01$ \\
DSM & $\bm{0.00 \pm 0.00}$ & $0.23 \pm 0.01$ & $0.33 \pm 0.02$ \\
\cmidrule(lr){1-1} \cmidrule(lr){2-4}
rand-RSM & $\bm{0.00 \pm 0.00}$ & $0.36 \pm 0.03$ & $0.40 \pm 0.01$ \\
CRJ-RSM & $0.35 \pm 0.02$ & $0.22 \pm 0.00$ & $0.08 \pm 0.01$ \\
LDN-RSM & $\bm{0.00 \pm 0.00}$ & $\bm{0.00 \pm 0.00}$ & $\bm{0.00 \pm 0.00}$ \\
\end{tabular}
\end{center}
\end{table}

\begin{table}
\caption{The mean absolute error ($\pm$ std.) on the test data for all models
for the language datasets. All results with mean and standard deviation below $10^{-2}$
are bold-faced.}
\label{tab:errors}
\begin{center}
\begin{scriptsize}
\begin{tabular}{lcccccc}
model & {Dyck1} & {Dyck2} & {Dyck3} & {$a^nb^n$} & {Palindrome} & {JSON} \\
\cmidrule(lr){1-1} \cmidrule(lr){2-7}
rand-ESN & $0.13 \pm 0.02$ & $0.23 \pm 0.05$ & $0.28 \pm 0.06$ & $\bm{0.00 \pm 0.00}$ & $\bm{0.00 \pm 0.00}$ & $\bm{0.01 \pm 0.00}$ \\
CRJ-ESN & $0.11 \pm 0.01$ & $0.12 \pm 0.02$ & $0.14 \pm 0.03$ & $\bm{0.00 \pm 0.00}$ & $\bm{0.01 \pm 0.00}$ & $0.01 \pm 0.00$ \\
LDN-ESN & $0.17 \pm 0.02$ & $0.23 \pm 0.02$ & $0.26 \pm 0.03$ & $\bm{0.00 \pm 0.00}$ & $0.07 \pm 0.01$ & $0.10 \pm 0.01$ \\
\cmidrule(lr){1-1} \cmidrule(lr){2-7}
GRU & $0.04 \pm 0.01$ & $0.05 \pm 0.01$ & $0.06 \pm 0.01$ & $\bm{0.00 \pm 0.00}$ & $\bm{0.00 \pm 0.00}$ & $\bm{0.00 \pm 0.00}$ \\
SRNN & $0.02 \pm 0.03$ & $\bm{0.00 \pm 0.00}$ & $0.03 \pm 0.03$ & $\bm{0.00 \pm 0.00}$ & $0.05 \pm 0.14$ & $0.08 \pm 0.16$ \\
DSM & $\bm{0.00 \pm 0.00}$ & $\bm{0.00 \pm 0.00}$ & $\bm{0.00 \pm 0.00}$ & $\bm{0.00 \pm 0.00}$ & $\bm{0.00 \pm 0.00}$ & $\bm{0.00 \pm 0.00}$ \\
\cmidrule(lr){1-1} \cmidrule(lr){2-7}
rand-RSM & $\bm{0.00 \pm 0.00}$ & $\bm{0.00 \pm 0.00}$ & $\bm{0.00 \pm 0.00}$ & $\bm{0.00 \pm 0.00}$ & $\bm{0.00 \pm 0.00}$ & $\bm{0.00 \pm 0.00}$ \\
CRJ-RSM & $\bm{0.00 \pm 0.00}$ & $\bm{0.00 \pm 0.00}$ & $\bm{0.00 \pm 0.00}$ & $\bm{0.00 \pm 0.00}$ & $\bm{0.00 \pm 0.00}$ & $\bm{0.00 \pm 0.00}$ \\
LDN-RSM & $\bm{0.00 \pm 0.00}$ & $\bm{0.00 \pm 0.00}$ & $\bm{0.00 \pm 0.00}$ & $\bm{0.00 \pm 0.00}$ & $\bm{0.00 \pm 0.00}$ & $\bm{0.00 \pm 0.00}$ \\
\end{tabular}
\end{scriptsize}
\end{center}
\end{table}

Tables~\ref{tab:benchmark_errors} and Table~\ref{tab:errors} report the mean average error across
data sets and models. We observe that the reservoir stack machine with
Lagrange delay reservoir (LDN-RSM) achieves almost zero error (bold-faced) on all datasets.
While other reservoirs succeed on the language tasks, they fail on the copy and repeat copy task.
This is to be expected because these tasks require a lossless reconstruction of past
states from the stack, which the Lagrange delay network is designed for \cite{LMU}.

Echo state networks without a stack fail almost all tasks, except for the
$a^nb^n$, palindrome, and JSON languages. This corresponds to the theoretical findings of
\citet{DMM}, indicating that ESNs cannot recognize general regular languages, let alone deterministic
context-free languages. The fact that some languages can still be solved is likely due to the fact
that all test sequences stay within the memory capacity of our model.

Interestingly, the deep models also fail for many tasks, especially the copy and repeat copy task.
This corresponds to prior findings of \citet{NTM_impl} that these tasks likely require tens of thousands
of unique sequences to be learned in a memory-augmented neural network, whereas we only present $100$
distinct sequences. Even for the language tasks, Stack-RNNs often fail, indicating that the correct
stack behavior is not easy to learn. Indeed, even the deep variation of our model (DSM), which receives
the same amount of training data as the reservoir stack machine, fails on the copy and repeat copy task, 
indicating that these tasks are not trivial to learn.

\begin{table}
\caption{The mean runtime ($\pm$ std.) in seconds as measured by Python's \texttt{time}
function for all models on the NTM benchmark datasets.}
\label{tab:benchmark_runtimes}
\begin{center}
\begin{tabular}{lccc}
model & {latch} & {copy} & {repeat copy} \\
\cmidrule(lr){1-1} \cmidrule(lr){2-4}
rand-ESN & $1.17 \pm 0.12$ & $0.48 \pm 0.01$ & $0.32 \pm 0.01$ \\
CRJ-ESN & $0.76 \pm 0.11$ & $0.11 \pm 0.01$ & $0.18 \pm 0.01$ \\
LDN-ESN & $1.71 \pm 0.19$ & $0.19 \pm 0.01$ & $0.23 \pm 0.01$ \\
\cmidrule(lr){1-1} \cmidrule(lr){2-4}
GRU & $97.1 \pm 4.1$ & $108.3 \pm 12.2$ & $191.5 \pm 74.4$ \\
SRNN & $295.0 \pm 12.4$ & $314.5 \pm 24.4$ & $536.3 \pm 113.5$ \\
DSM & $619.3 \pm 20.3$ & $548.9 \pm 30.3$ & $1045.4 \pm 211.6$ \\
\cmidrule(lr){1-1} \cmidrule(lr){2-4}
rand-RSM & $10.25 \pm 0.62$ & $5.20 \pm 0.14$ & $9.61 \pm 0.65$ \\
CRJ-RSM & $51.27 \pm 5.16$ & $4.11 \pm 0.14$ & $7.78 \pm 0.38$ \\
LDN-RSM & $26.74 \pm 2.04$ & $3.36 \pm 0.13$ & $10.25 \pm 0.46$ \\
\end{tabular}
\end{center}
\end{table}

\begin{table}
\caption{The mean runtime ($\pm$ std.) in seconds as measured by Python's \texttt{time}
function for all models on the language datasets.}
\label{tab:runtimes}
\begin{center}
\begin{scriptsize}
\begin{tabular}{lcccccc}
model & {Dyck1} & {Dyck2} & {Dyck3} & {$a^nb^n$} & {Palindrome} & {JSON} \\
\cmidrule(lr){1-1} \cmidrule(lr){2-7}
rand-ESN & $1.68 \pm 0.30$ & $1.78 \pm 0.13$ & $1.93 \pm 0.40$ & $0.89 \pm 0.04$ & $1.29 \pm 0.29$ & $1.01 \pm 0.06$ \\
CRJ-ESN & $1.27 \pm 0.32$ & $0.88 \pm 0.13$ & $0.98 \pm 0.21$ & $0.38 \pm 0.02$ & $0.61 \pm 0.16$ & $0.80 \pm 0.05$ \\
LDN-ESN & $2.27 \pm 0.29$ & $1.28 \pm 0.10$ & $1.22 \pm 0.26$ & $0.92 \pm 0.12$ & $1.04 \pm 0.32$ & $0.95 \pm 0.15$ \\
\cmidrule(lr){1-1} \cmidrule(lr){2-7}
GRU & $45.6 \pm 3.1$ & $45.5 \pm 3.0$ & $42.3 \pm 2.0$ & $52.6 \pm 8.9$ & $47.9 \pm 2.5$ & $37.8 \pm 1.8$ \\
SRNN & $114.8 \pm 10.7$ & $114.0 \pm 10.3$ & $102.5 \pm 6.2$ & $131.0 \pm 24.7$ & $122.0 \pm 9.5$ & $88.0 \pm 5.9$ \\
DSM & $285.0 \pm 31.3$ & $282.1 \pm 30.8$ & $249.1 \pm 17.7$ & $270.3 \pm 37.5$ & $290.3 \pm 27.0$ & $217.6 \pm 17.3$ \\
\cmidrule(lr){1-1} \cmidrule(lr){2-7}
rand-RSM & $11.52 \pm 0.72$ & $16.64 \pm 1.40$ & $17.80 \pm 1.78$ & $4.01 \pm 0.08$ & $6.09 \pm 0.42$ & $13.16 \pm 1.18$ \\
CRJ-RSM & $11.08 \pm 0.69$ & $15.06 \pm 1.14$ & $17.08 \pm 1.61$ & $4.15 \pm 0.07$ & $7.06 \pm 0.50$ & $11.42 \pm 0.99$ \\
LDN-RSM & $13.71 \pm 0.97$ & $14.34 \pm 1.40$ & $14.22 \pm 0.84$ & $5.06 \pm 0.22$ & $6.49 \pm 0.42$ & $11.90 \pm 0.79$ \\
\end{tabular}
\end{scriptsize}
\end{center}
\end{table}

Tables~\ref{tab:benchmark_runtimes} and~\ref{tab:runtimes} show the runtime needed for training and evaluating all models on all datasets. Unsurprisingly, RSMs take more time compared to ESNs due to the
stack mechanism and because we need to train four classifiers instead of one linear regression,
yielding a factor of $\approx 10$ for LDN-RSMs. GRUs are about seven times slower compared to an LDN-RSM, SRNNs are about $20$ times slower, and DSMs about $35$ times slower.

Overall, we conclude that reservoir stack machines can solve tasks that are
impossible to solve for ESNs and hard to solve for deep networks. Additionally, while RSMs are much
slower compared to pure echo state networks, they are still much faster compared to deep networks
(even for small training data sets).

\section{Conclusion}

In this paper, we presented the reservoir stack machine (RSM), a combination of an echo state network
with a stack. We have shown that a sufficiently rich reservoir suffices to simulate any
LR(1)-automaton, whereas a constant-sized memory only suffices for finite state automata.
We have evaluated our model on three benchmark tasks for Neural Turing Machines (latch, copy, and repeat copy)
and six deterministic context-free languages. ESNs struggled with all three benchmark tasks and most LR(1) languages
and even deep models were unable to solve the copy and repeat copy task.

By contrast, RSMs could solve all tasks with zero generalization error.
For the LR(1) languages, this is independent of the choice of reservoir, whereas a Legendre delay
network was required for the copy and repeat copy task. The Legendre network has the advantage
that it can provably decode past inputs via a linear operator, which simplifies the output function
on these tasks.

We admit that a crucial limitation of RSMs is that they require additional training data in the form of
desired pop, push, and shift behavior. However, RSMs can generalize this behavior from examples
to longer sequences, indicating that actual learning takes place. Further, even with this additional
training data, a deep learning model failed to solve tasks that an LDN-RSM could solve, indicating that
the learning task is still non-trivial.
Accordingly, we conclude that RSMs provide a novel way to learn difficult stack behavior
within seconds and with few short reference sequences.

Future research could extend the reservoir stack machine with a second stack to a full Turing
machine and try to find mechanisms in order to construct desired pop, push, and shift behavior
for training data. Even in the present form, though, we hope that the reservoir stack machine
provides an interesting new avenue to explore memory-augmented neural networks in a way that is
faster and more reliable to train.

\section{Acknowledgements}

Funding by the German Research Foundation (DFG) under grant
numbers PA 3460/1-1 and PA 3460/2-1 is gratefully acknowledged.

\bibliographystyle{elsarticle-harv}
\bibliography{literature}

\begin{appendix}
\section{Additional Experiments}

To verify that the worse results for pure echo state networks are not just due to less parameters,
we repeated all experiments with a reservoir with $1024$ neurons, i.e.\ four times the number as for
the RSM models to counteract the four output functions that the RSM model has available. The errors
are shown in Table~\ref{tab:errors_1024}. Qualitatively, we observe the same results as in 
Tables~\ref{tab:benchmark_errors} and~\ref{tab:errors}, namely that the ESN models are unable to solve
latch, copy, or repeat copy, and none of the Dyck languages, but that some reservoirs achieve zero
error on $a^nb$, palindromes, and JSON.

\begin{table}
\caption{The mean absolute error ($\pm$ std.) on the test data for ESN models with
$1024$ neurons. All results with mean and standard deviation below $10^{-2}$
are bold-faced.}
\label{tab:errors_1024}
\begin{center}
\begin{tabular}{lccc}
dataset & {rand-ESN} & {CRJ-ESN} & {LDN-ESN} \\
\cmidrule(lr){1-1} \cmidrule(lr){2-4}
latch & $0.42 \pm 0.03$ & $0.39 \pm 0.01$ & $0.43 \pm 0.01$ \\
copy & $0.24 \pm 0.01$ & $0.21 \pm 0.00$ & $0.22 \pm 0.00$ \\
repeat copy & $0.53 \pm 0.04$ & $0.33 \pm 0.01$ & $0.31 \pm 0.01$ \\
Dyck1 & $0.14 \pm 0.02$ & $0.14 \pm 0.02$ & $0.18 \pm 0.02$ \\
Dyck2 & $0.58 \pm 0.64$ & $0.15 \pm 0.02$ & $0.23 \pm 0.02$ \\
Dyck3 & $0.34 \pm 0.08$ & $0.16 \pm 0.03$ & $0.27 \pm 0.03$ \\
$a^nb^n$ & $\bm{0.00 \pm 0.00}$ & $\bm{0.00 \pm 0.00}$ & $\bm{0.00 \pm 0.00}$ \\
Palindrome & $\bm{0.00 \pm 0.00}$ & $\bm{0.00 \pm 0.00}$ & $0.07 \pm 0.01$ \\
JSON & $\bm{0.00 \pm 0.00}$ & $\bm{0.00 \pm 0.00}$ & $0.10 \pm 0.01$ \\
\end{tabular}
\end{center}
\end{table}

The runtimes are shown in Table~\ref{tab:runtimes_2014}. As to be expected, we observe
longer runtimes than in Tables~\ref{tab:benchmark_runtimes} and~\ref{tab:runtimes}
because the reservoirs are much larger. As before, the CRJ is the fastest reservoir
overall and scales roughly linearly with the reservoir size, whereas the random and
Legendre reservoirs appear to scale worse than linear, approaching and sometimes exceeding
the runtime of the reservoir stack machine models with $256$ neurons.

\begin{table}
\caption{The mean runtime ($\pm$ std.) in seconds as measured by Python's \texttt{time}
function for ESN models with $1024$ neurons.}
\label{tab:runtimes_2014}
\begin{center}
\begin{tabular}{lccc}
dataset & {rand-ESN} & {CRJ-ESN} & {LDN-ESN} \\
\cmidrule(lr){1-1} \cmidrule(lr){2-4}
latch & $6.00 \pm 0.22$ & $1.43 \pm 0.06$ & $12.85 \pm 0.44$ \\
copy & $0.93 \pm 0.04$ & $0.20 \pm 0.03$ & $0.38 \pm 0.03$ \\
repeat copy & $20.88 \pm 1.35$ & $1.61 \pm 0.22$ & $3.83 \pm 0.18$ \\
Dyck1 & $15.34 \pm 0.67$ & $1.71 \pm 0.10$ & $13.33 \pm 0.84$ \\
Dyck2 & $15.66 \pm 0.53$ & $1.81 \pm 0.15$ & $8.15 \pm 0.41$ \\
Dyck3 & $9.28 \pm 0.55$ & $1.76 \pm 0.11$ & $6.29 \pm 0.54$ \\
$a^nb^n$ & $10.54 \pm 0.10$ & $0.89 \pm 0.01$ & $8.73 \pm 0.08$ \\
Palindrome & $11.28 \pm 0.82$ & $1.15 \pm 0.31$ & $6.48 \pm 0.42$ \\
JSON & $6.59 \pm 0.63$ & $1.78 \pm 0.26$ & $4.03 \pm 0.26$ \\
\end{tabular}
\end{center}
\end{table}

\end{appendix}

\end{document}